%% file: main.tex
\documentclass[letterpaper]{article} 
\usepackage{aaai25}  
\usepackage{times}  
\usepackage{helvet}  
\usepackage{courier}  
\usepackage[hyphens]{url}  
\usepackage{graphicx} 
\usepackage{natbib}  
\usepackage{caption} 
\usepackage{algorithm}
\usepackage{algorithmic}

\usepackage{latexsym}
\usepackage{amssymb}
\usepackage{amsmath}
\usepackage{amsthm}
\usepackage{booktabs}
\usepackage{enumitem}
\usepackage{comment}
\usepackage{multirow}
\usepackage{arydshln}
\usepackage{xcolor}
\usepackage{xspace}
\usepackage{subcaption}

\newcommand{\resourcename}{\textsc{PDDL2NL}}
\newcommand{\PLLM}{P-LLM}
\newcommand{\TLLM}{T-LLM}
\newcommand{\APBLLM}{2NL-LLM}

\newcommand{\Pex}{few-shot example}

\newcommand{\appdom}{Appendix~B}
\newcommand{\appapb}{Appendix~C}
\newcommand{\appprompts}{Appendix~D}
\newcommand{\appcost}{Appendix~E}
\newcommand{\appreact}{Appendix~F}
\newcommand{\exBasic}{$\mathtt{Bas}$\xspace}
\newcommand{\exCoT}{$\mathtt{CoT}$\xspace}
\newcommand{\exReAct}{$\mathtt{ReA}$\xspace}
\newcommand{\exAct}{$\mathtt{Act}$\xspace}
\newcommand{\exBFS}{$\mathtt{BrFS}$\xspace}
\newcommand{\exLMC}{$\mathtt{lmc}$\xspace}
\newcommand{\exFF}{$\mathtt{ff}$\xspace}
\newcommand{\exRnd}{$\mathtt{rnd}$\xspace}

\newcommand{\exManual}{$\mathtt{Valm23}$\xspace}
\newcommand{\exPDDL}{$\mathtt{PDDL}$\xspace}
\newcommand{\exTemplate}{$\mathtt{Tpl}$\xspace}

\usepackage{newfloat}
\usepackage{listings}
\DeclareCaptionStyle{ruled}{labelfont=normalfont,labelsep=colon,strut=off} 
\floatstyle{ruled}
\newfloat{listing}{tb}{lst}{}
\floatname{listing}{Listing}
\title{Automating the Generation of Prompts\\ for LLM-based Action Choice in PDDL Planning}
\author {
    Katharina Stein\textsuperscript{\rm 1},
    Daniel Fi\v{s}er\textsuperscript{\rm 3},
    J\"org Hoffmann\textsuperscript{\rm 1,2},
    Alexander Koller\textsuperscript{\rm 1}
}
\affiliations {
    \textsuperscript{\rm 1}Saarland Informatics Campus, Saarland University, Saarbr\"ucken Germany\\
    \textsuperscript{\rm 2}German Research Center for Artificial Intelligence (DFKI), Saarbr\"ucken, Germany\\
    \textsuperscript{\rm 3}Aalborg University, Denmark\\
    \{kstein,koller\}@lst.uni-saarland.de, danfis@danfis.cz, hoffmann@cs.uni-saarland.de
}

\begin{document}

\maketitle

\begin{abstract}
Large language models (LLMs) have revolutionized a large variety of
NLP tasks. An active debate is to what extent they can do reasoning
and planning. Prior work has assessed the latter in the specific
context of PDDL planning, based on manually
converting three PDDL domains into natural language (NL) prompts. Here we automate this
conversion step, showing how to leverage an LLM to automatically generate NL
prompts from PDDL input. Our automatically
generated NL prompts result in similar LLM-planning performance as the
previous manually generated ones. Beyond this, the automation
enables us to run much larger experiments, providing for the first
time a broad evaluation of LLM planning performance in
PDDL.
%
%
%
Our NL prompts yield better performance than PDDL prompts and simple template-based NL prompts. Compared to symbolic planners, LLM planning lags far behind; but in some domains, our best LLM
configuration scales up further than A$^\star$ using LM-cut.
\end{abstract}

%
\begin{links}
     \link{Code and datasets}{https://github.com/minecraft-saar/autoplanbench}
 \end{links}

\section{Introduction}\label{sec:intro}



Large language models (LLMs) have revolutionized a large variety of
natural language processing tasks. A recent research trend
investigates whether LLMs can also do planning. The word ``planning''
here is used in a broad sense, encompassing, for example, robot
control \cite{saycan2022i}, text-based games \cite{yao2022react} or
Minecraft \cite{wang2023describe, zhu2023ghost}, but
also less structured tasks such as question answering
\cite[e.g.,][]{NEURIPS2022_9d560961, khot-etal-2021-text}, visual
programming \cite{Gupta_2023_CVPR}, and the orchestration of API calls
\cite{prasad2023adapt}.


Here we address PDDL planning. The use of LLMs in this context is, at this stage,
still in its infancy. First works explored what form of input to
provide to the LLM (PDDL, natural language)
\cite{silver2022pddl,valmeekam2023planbench,valmeekam2023planning};
recent work explored the generation of program code for generalized
planning \cite{silver2024generalized}. Here, we follow up on the
prominent work line by \citet{valmeekam2023planbench,valmeekam2023planning}, who
investigated the ability of LLMs to produce plans for PDDL planning tasks. Valmeekam et
al.\ experiment with three wide-spread benchmark domains, Blocksworld, Depots and Logistics, for each of which
they manually engineer natural language (NL) descriptions of the actions
and predicates. They ask the LLM to produce a plan, thus serving as a
form of planner that gives no plan correctness (nor
optimality) guarantee. Valmeekam et al.\ find that LLMs (both GPT-3.5
and GPT-4) are unable to reliably produce correct plans in their 3
benchmark domains, lagging far behind symbolic planning methods.

%



We extend Valmeekam et al.'s work by automating the conversion of PDDL
into NL prompts for LLM plan generation (and, more
generally, action choice mechanisms, see below). We thus turn the use of LLMs
with NL prompts into domain-independent planning machinery that
does not rely on any input other than the PDDL itself. A key 
challenge here is to ensure that the (syntactic and semantic) relationship
between an action and its arguments is captured correctly, and that
PDDL object types are made explicit in the action descriptions in an
adequate way.

%
We address this by leveraging LLMs themselves to support
the conversion from PDDL to NL. We first use GPT-4o
\cite{gpt-4o} to convert PDDL predicates into NL snippets
based on a few generic conversion examples. We then generate NL
snippets for PDDL action schemas, by first converting the
preconditions and effects based on a simple composition of the
previously generated predicate snippets, and then doing the final
conversion step with an LLM, based again on a few generic examples.
Lastly, we compose the final NL description of the PDDL task for the LLM
prompt from predicate and action snippets together with NL descriptions of
initial state and goal.
Our conversion approach ensures that, structurally, all predicate and actions schemas are preserved correctly in the NL descriptions.
Experimenting with the same benchmarks used by Valmeekam et al., we
find that our automatically generated prompts result in
similar performance as the previous manually generated
ones.



Beyond this, we provide the first
broad evaluation of LLM performance in PDDL planning, comparing our automatically-generated NL prompts to PDDL prompts and simple template-based prompts, comparing four variants of LLM action choice
mechanisms\footnote{We use ``LLM action choice mechanism''
here as a generic term encompassing both plan generation and action
policies, and to emphasize that (in difference to symbolic planners)
these mechanisms do not use any search, instead choosing actions
directly.}, and comparing LLM planning to several symbolic planning baselines, on a broad range of benchmark domains. 
%

Specifically, we experiment with:
\begin{itemize}
\item \textbf{Basic LLM Planning:} Domain and problem description
  provided once, LLM generates a plan. Same as \citet{valmeekam2023planbench,valmeekam2023planning}.
\item \textbf{CoT LLM Planning:} The LLM also generates a plan,
  but the prompt is enriched by Chain-of-Thought (CoT) prompting
  \cite{NEURIPS2022_9d560961}, asking the LLM to generate
  ``thoughts'' between the predicted actions in the plan.
  \citet{valmeekam2023planning} experimented with a simple version
  of this, producing the states before and after each
  action; here we allow more flexible reasoning, e.g., about the next
  required actions.
\item \textbf{Act:} Here the LLM is used as an action policy instead
  of a plan generator, choosing an individual action for the state at
  each step in the plan. (A similar idea was explored by
  \citet{yao2022react} in non-PDDL-planning contexts.)
  %
  %
\item \textbf{ReAct:} Inspired by
  \citet{yao2022react}; uses CoT within Act, generating intermediate
  thoughts between actions.
\end{itemize}
Our primary experiment is on 13 domains for which we generate small instances, targeting the relevant scaling range. For a subset of configurations, we also run experiments on all IPC domains, with their original instances, to the extent feasible (monetary cost for LLM calls is the main issue here). We provide
comparisons to (1) LLM action choice based on PDDL prompts
\cite{valmeekam2023planning,silver2022pddl} rather than
NL encodings thereof; (2) a simple template-based method to convert PDDL into NL prompts; (3) random action selection as a
sanity test; (4) blind breadth-first search as a trivial symbolic
baseline; as well as (5) strong optimal and
satisficing planner baselines \cite{lmcut, hoffmann2001ff}.


Compared to the alternative automatic prompting methods (1) and (2), our NL prompts
often yield better performance, in particular as they allow to incorporate Chain-of-Thought (in CoT and ReAct). All four LLM action
choice mechanisms soundly beat (3) random action choice,
showing that the LLM does carry
\emph{some} information about general PDDL planning (not self-evident given the paucity of internet text about most planning
benchmark domains).\footnote{A similar observation was made by \citet{silver2022pddl},
but in a more limited setting considering PDDL prompts and a set-up
comparable to our Basic LLM Planning configuration, which exhibits
much worse performance than our strongest methods.}
The comparisons (4) and (5) to symbolic planners are
less favorable. The satisficing planner (directly
comparable as it does not provide a plan-quality guarantee) reigns
supreme throughout. On the positive side, ReAct outperforms
breadth-first search in 7 domains and the optimal planner in 6 domains.
While these are isolated islands of good performance, they do show
promise for LLM planning abilities, in particular as this performance
is obtained without any search.

\section{Background}\label{sec:rel_work}

PDDL \cite{pddl1998ghallab} has been established in the context of the
International Planning
Competitions\footnote{\url{https://www.icaps-conference.org/competitions/}}.
A planning task in PDDL consists of a domain and problem file.
The domain file defines the world model using predicates describing
possible world states, and actions whose execution changes the current state.
The problem file defines a specific instance from the domain by specifying
available objects, the initial state and the goal.

Each action is defined by its precondition specifying what has to be true
in the state where the action is applied, and by its effect saying what
will become true (add effect) and false (delete effect) in the resulting
state.
The solution to a planning problem is a plan---a sequence of actions
leading from the initial state to a state where the goal condition holds.

\begin{figure*}[t]
    \centering
    \subfloat[PDDL domain definition with the type hierarchy ($\mathcal{T}$), predicates ($\mathcal{P}$), and the ``drive-truck'' action ($\mathcal{A}$).]{\includegraphics[width=0.35\textwidth]{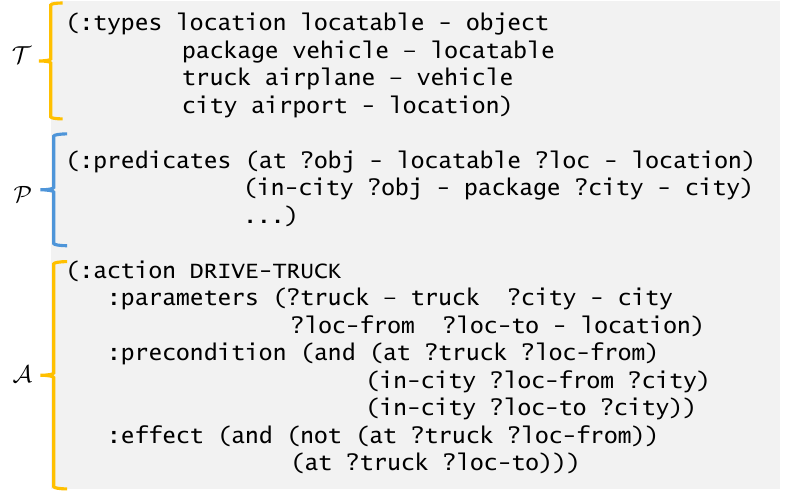}\label{fig:pddl_domain}}
    \qquad
    \subfloat[NL domain description consisting of the available actions with parameters ($\mathcal{A^{PA}}$), their preconditions ($\mathcal{A^{PR}}$) and effects ($\mathcal{A^E}$) and the type hierarchy ($\mathcal{T}$).]{\includegraphics[width=0.55\textwidth]{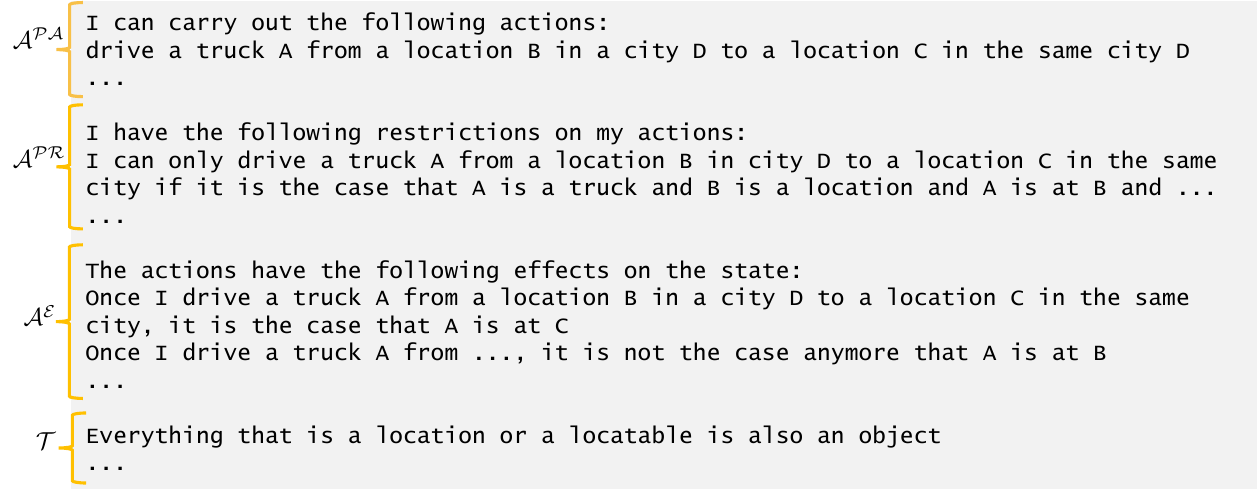}\label{fig:nl_domain}}
    
    \subfloat[PDDL problem file stating the available objects with types ($\mathcal{O}$), the initial state ($\mathcal{I}$) and the goal condition ($\mathcal{G}$).]{\includegraphics[width=0.35\textwidth]{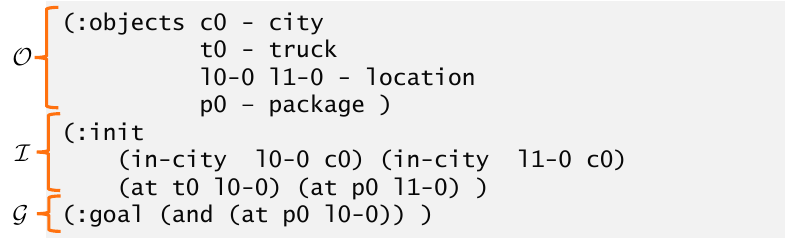}\label{fig:pddl_prob}}
    \qquad
    \subfloat[NL problem description stating the goal ($\mathcal{G}$), the available objects ($\mathcal{O}$) and the initial state ($\mathcal{I}$).]{\includegraphics[width=0.55\textwidth]{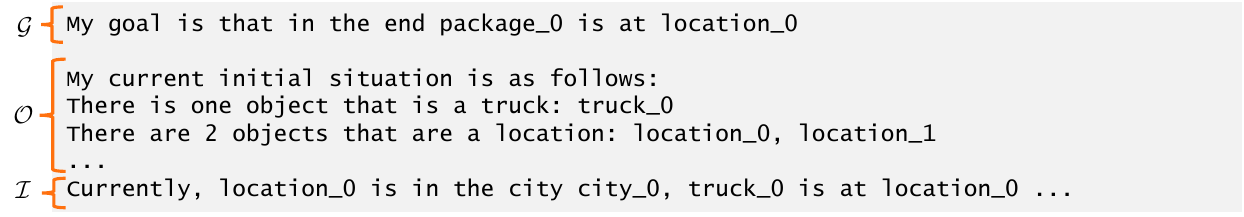}\label{fig:nl_problem}}
    \caption{\label{fig:domain_ex}%
Part of the Logistics PDDL domain 
and problem 
file and the NL descriptions generated by {\resourcename}. 
}
\end{figure*}

Figure~\ref{fig:pddl_domain} shows an excerpt from the Logistics benchmark domain
that models delivering packages with trucks within cities and with
planes between cities. 
The action ``drive-truck'' describing driving a truck between two locations
is parametrized with variables
``?truck'', ``?loc-from'', ``?loc-to'' and ``?city'' whose
instantiation with objects specifies the truck being driven, the start and
the destination location and the city in which the locations are.
The precondition states that the action can only be executed if the truck
is at location ?loc-from and both locations are in city ?city.
The effect makes the atom (at ?truck ?loc-from) false and (at ?truck
?loc-to) becomes true, i.e., it changes the state by moving the truck
?truck from ?loc-from to ?loc-to.

Variables can also have types restricting which objects can be used for
their instantiation.
For example, ?truck has the type ``truck'' which in our example
has only one corresponding object ``t0''.
There are different variants of the PDDL language with varying
expressiveness.
Here, we consider a PDDL subset allowing variable typing and
restricted to conjunctive conditions with negation.

\paragraph{PDDL and LLMs.}
\citet{valmeekam2023planbench} introduced Planbench, a benchmark framework for assessing different aspects of reasoning capabilities of LLMs based on classical planning problems formulated in PDDL. Their assessment pipeline can take PDDL domain and problem files as input as well as natural language (NL) descriptions of the PDDL domain and problem files for assessing LLMs on NL problem formulations. 
For the NL inputs, they manually create a NL description of the domain file and handcraft NL translations for the individual PDDL actions, predicates and for object names. The individual translations are used to compose NL descriptions of problem files and plans. 

In their assessment pipeline, \citet{valmeekam2023planbench} prompt an LLM to generate a plan based on the NL descriptions of the domain, the initial and goal state and few-shot examples, i.e., example problems with their corresponding plans for in-context learning. The predicted NL action sequences get translated back into PDDL by a domain-dependent translator  and are automatically evaluated by a plan validator. This process allows a systematic and objective evaluation of the performance of LLMs on different reasoning-related test cases.
However, the approach includes several manually provided per-domain components. 

\citet{silver2022pddl} assess the planning capabilities of LLMs when using prompts that do not contain any natural language and consist of the target problem definition and two few-shot examples in PDDL. They evaluate OpenAI's Codex LLM \cite{codex}, an LLM specifically trained to generate code for NL inputs, and find that in some domains such as Gripper and Movie, LLMs can solve even large problems while they completely fail on more than half of the evaluated domains, including Blocksworld. \citet{valmeekam2023planning} also assess the capabilities of pretrained LLMs on PDDL but their prompts include a general task description in NL and the PDDL domain in addition to the target problem and examples.

There is a lot of work being done on LLMs in the context of PDDL planning most of which is only remotely related to ours. For example, \citet{Rossetti_2024} investigate learning per-domain generalizing policies by training transformer models on PDDL from scratch.
Another line of research focuses on using LLMs to guide symbolic search \cite[e.g.][]{hazra24saycanpay, hao-etal-2023-reasoning, zhou24language_agent_tree_search}.   
\citet{katz24thought_of_search} and \citet{silver2024generalized} propose to let LLMs generate Python code to be used for deriving plans. 

\begin{figure}[t]
    \centering
    \includegraphics[width=0.47\textwidth]{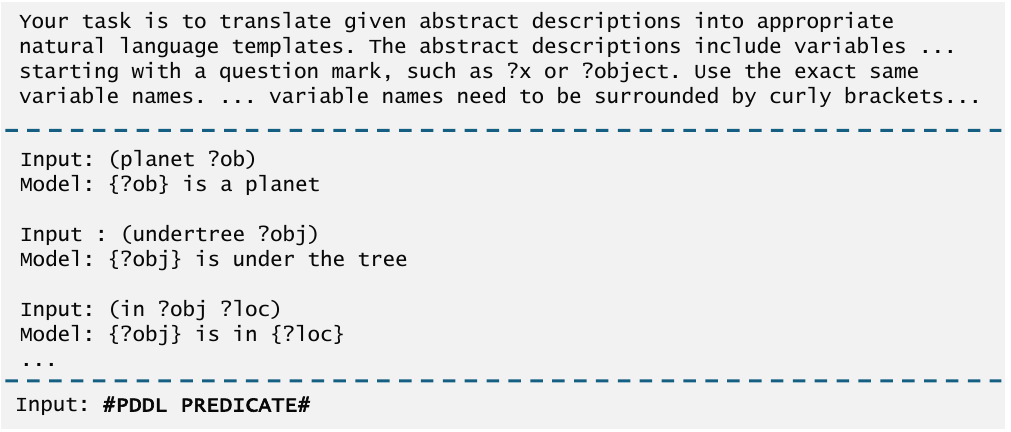}
    \caption{Part of the prompt for converting PDDL predicates into NL consisting of the task description (top), few-shot examples (middle) and the target predicate (bottom). }
    \label{fig:conv_prompt}
\end{figure}

\section{Converting PDDL into Natural Language}\label{sec:method}

We contribute an automatic conversion, \resourcename, of PDDL domains and problems
into NL descriptions.
Our method first produces PDDL-to-NL mappings for objects, predicates, and actions, then joins these together into a prompt tasking the LLM to
solve the described planning instance. To achieve this automatic conversion, we leverage an LLM
(called \APBLLM, short for \resourcename-LLM), with few-shot
prompting on generic examples (fixed once and used for any input PDDL)  for each part of the conversion.

Figure \ref{fig:conv_prompt} illustrates the input prompt received by the
{\APBLLM} for translating PDDL predicates into NL.
The prompt consists of three parts.
The first part is a manually designed instruction explaining the PDDL-to-NL translation.
The second part consists of several examples illustrating to the {\APBLLM} how
to conduct the translation.
These so-called few-shot examples are hand-crafted, but we use exactly the
same examples independently of the target PDDL domain.
We use seven examples of predicates of arity ranging from 0 to 2.
After the examples, we pass the actual predicate that we want to translate\footnote{We use \#PART\# to mark parts of the shown prompts that
are placeholders for actual content omitted for the presentation throughout
the paper.} to the LLM, 
i.e., we tell the {\APBLLM} which PDDL predicate we want to
translate to NL (e.g., we provide ``(at ?x ?y)'' as input)
and expect the {\APBLLM} to return its NL description (e.g.,
``\{?x\} is at \{?y\}''). 
Table~\ref{tab:pddl2text} (top) shows the NL descriptions of predicates
generated in the Logistics domain.
Note that the NL descriptions are constructed as \textit{snippets} with
placeholders for actual objects (e.g., ``\{?obj\}'') which are replaced later
when constructing the final NL task description.
Therefore, we require the LLM to generate snippets that include all variable names from the PDDL, each surrounded by brackets. We automatically check whether this condition is met and if not the \APBLLM\ is prompted once to revise the previous output. 

NL snippets of actions are generated analogously using a similar
prompt with different examples.
In this case, each example consists of the name of the action, its parameters, and NL descriptions of the
preconditions and effects that are constructed using the NL snippets of the predicates generated by the \APBLLM\ as described above.
The precondition is constructed by joining NL descriptions of
its positive atoms by ``and'' and the conjoined negative preconditions are preceded by ``it is not the case that''. The NL descriptions of the positive (add) and negative (delete) effects are conjoined analogously.
Moreover, we compile away parameter types using unary predicates
\cite[e.g.,][]{helmert2009concise}. 
We use the same four hand-crafted few-shot examples of actions independently of the target domain. As for predicates, snippets not matching the formal requirements are revised once. 

\begin{table}[t]
    \centering
{\def\arraystretch{1}\tabcolsep=2.3pt
    \small
    \begin{tabular}{|l|l|l|}
    \hline
         In: & (truck ?truck ) & (location ?location) \\
         \hdashline
         Out: & \{?truck\} is a truck & \{?location\} is a location \\
         \hline
         In: & (at ?obj ?loc) & (in-city ?obj ?city)\\
         \hdashline
         Out: & \{?obj\} is at \{?loc\} & \{?obj\} is in the \{?city\}\\
         \hline\hline
         In: & \multicolumn{2}{|l|}{action: drive-truck} \\
         & \multicolumn{2}{|l|}{parameters: (?truck ?l-from ?l-to ?city)}\\
         & \multicolumn{2}{|l|}{preconditions of drive-truck: ?truck is a truck and } \\
         & \multicolumn{2}{|l|}{\qquad ?l-from is a location and ?l-to is a location and  } \\
         & \multicolumn{2}{|l|}{\qquad ?city is a city and ?truck is at ?l-from and ?l-from} \\
         & \multicolumn{2}{|l|}{\qquad  is in city ?city and ?l-to is in city ?city} \\
         & \multicolumn{2}{|l|}{effects of drive-truck: it becomes true that ?truck is at ?l-to} \\
         & \multicolumn{2}{|l|}{\qquad and it is not the case anymore that ?truck is at ?l-from} \\
         \hdashline
         Out: & \multicolumn{2}{|l|}{drive truck \{?truck\} from location \{?l-from\} in city }\\
                 & \multicolumn{2}{|l|}{\qquad \{?city\} to location \{?l-to\} in the same city}\\
    \hline
    \end{tabular}
    
}
\caption{\label{tab:pddl2text}%
Example PDDL-to-NL translation by the {\APBLLM} in the Logistics domain;
predicates at the top; the ``drive-truck'' action at the bottom.}
\end{table}

Table \ref{tab:pddl2text} (bottom) shows an example translation of the
``drive-truck'' action from the Logistics domain.
It illustrates two interesting characteristics of our NL snippets.
First, the
order of the arguments in the generated NL snippet can deviate from the
order of the parameters in the input PDDL domain.
The order of parameters can be
arbitrary and might not match a natural sounding or even syntactically
correct order of arguments of the action verb in NL. We therefore include
one few-shot example where the order deviates in the prompt for the
\APBLLM\ to prevent the LLM from inferring that the order needs to be
identical.
Second, the generated NL snippet of ``drive-truck" states the type of
each parameter, i.e., it makes use of the information
from preconditions to infer appropriate types. The complete conversion
prompts can be found in \appapb.

With the NL descriptions of predicates and actions
in the form of snippets, we proceed with the generation of the domain
and problem NL descriptions of the input PDDL task.

Figure \ref{fig:nl_domain} shows an excerpt from the NL \emph{domain} description of the
PDDL Logistics domain (Figure~\ref{fig:pddl_domain}).
NL domain descriptions are designed to include the same information as the
input PDDL.
They start with the description of all possible actions ($\mathcal{A^{PA}}$),
followed by their preconditions ($\mathcal{A^{PR}}$) and effects
($\mathcal{A^{E}}$). If the domain is typed, a verbalization of the type
hierarchy is added also ($\mathcal{T}$). Our template takes care
of the statements introducing each part of the prompt (e.g. ``I can carry
out the following actions:'') as well as of adequately composing the
preconditions and effects into NL sentences (e.g. ``Once I \#ACTION\# it is
not the case anymore that \#EFFECT\#'' for delete effects). The positive
and negative preconditions are presented in two individual sentences. The
same applies to the add and delete effects.
Moreover, we use a heuristic to add
indefinite determiners to ensure that the domain encodings refer to objects
in general instead of specific objects (e.g. ``drive \textbf{a} truck A'')
and that the referring expressions allow to correctly infer which
expressions refer to the same object.

The NL \emph{problem} descriptions (see example
in~Figure~\ref{fig:pddl_prob} and \ref{fig:nl_problem})
specify the goal condition ($\mathcal{G}$),
available objects with their types ($\mathcal{O}$), and the
initial state ($\mathcal{I}$).

The object names in PDDL problems can be any strings of characters and they
often consist of single letters and numbers. For the NL description of the
planning problems, more natural and semantically related object names are desirable. 
Our method generates new object names based on their types. If a domain is typed, we name each object after its (most specific) type and enumerate them, e.g. ``t0'' (Figure \ref{fig:pddl_prob}) becomes ``truck\_0'' (Figure \ref{fig:nl_problem}). Otherwise, we use the most general PDDL type ``object'' for all object names.

The NL descriptions of the initial state and goal are constructed using the
NL descriptions of the corresponding predicates
obtained by the {\APBLLM}.
For example, the goal ``(at p0 l0-0)'' (Figure \ref{fig:pddl_prob}) is converted into ``package\_0 is at
location\_0'' and appended to ``My goal is that in the end'' (Figure
\ref{fig:nl_problem}). If there is more than one goal fact, they are
conjoined using ``and''. The description of the initial state is
constructed analogously but starts with ``Currently,''. 

Our approach of letting the LLM generate NL snippets for the predicates and actions that are then used to compose the domain and problem descriptions in a rule-based way ensures that the logical structure of the PDDL definition is preserved. Therefore, 
our method guarantees that all action schemas are structurally correct in the NL description.


\section{LLM Action Choice Mechanisms}
\label{sec:mechanisms}


\begin{figure}[t]
    \centering
    \includegraphics[width=0.47\textwidth]{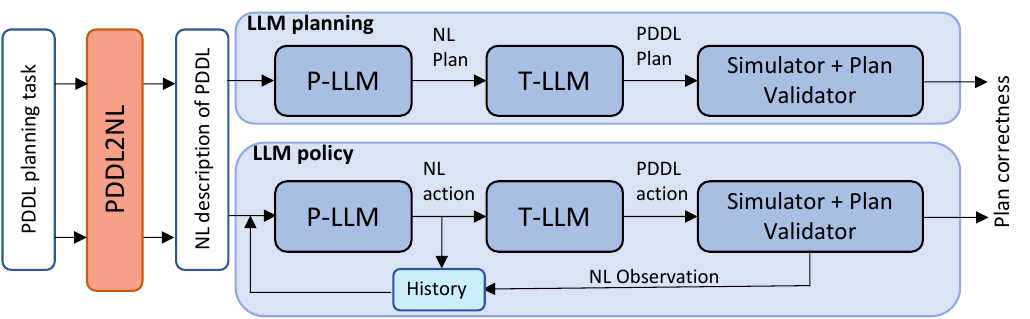}
    \caption{Overview of the set-up for the LLM plan generation and LLM action policy usage.
}
    \label{fig:approach_overview}
\end{figure}

Our automatic PDDL-to-NL translation can be used in concert with
different LLM action-choice techniques.  We distinguish LLM
\emph{planning} techniques (returning a whole action sequence at once)
and LLM \emph{policy} techniques (returning one action at a time).
They both consist of three core components, namely {\PLLM},
{\TLLM} and a simulator (see Figure~\ref{fig:approach_overview}).  The
NL description of the input PDDL task constructed using
{\resourcename} described in Section~\ref{sec:method} is passed to the
LLM (denoted as {\PLLM}) responsible for the actual action choice
(i.e., either a sequence of actions or a single action).  The output
of the {\PLLM} is in the form of NL.  So, we pass its output to
another LLM (denoted {\TLLM}) which translates the NL descriptions of
actions (or action sequences) back to the PDDL format.  Finally, we
pass the PDDL actions returned by the {\TLLM} to the simulator
responsible for validation and analysis of the system's action choice.
We describe each component below.

\paragraph{Simulator.} 
We implemented an environment simulator for PDDL tasks.
The simulator makes use of the plan validator
VAL\footnote{\url{https://github.com/KCL-Planning/VAL}} to determine
whether an action is applicable in the current state, or to obtain the
unsatisfied preconditions if it is not applicable.  
For LLM planning, the simulator only validates the output action
sequence and determines whether it solves the task.
For LLM policies, the simulator updates the world state after every
action chosen by the LLM. 

Instead of giving back the resulting world state in PDDL format,
taking inspiration from prior work on non-PDDL forms of planning with
LLMs \cite{wang2022scienceworld,shridhar2021alfworld,yao2022react,
  lin2023swiftsage}, our simulator produces a NL observation about
actions' effects as feedback for the next LLM-policy step.  
If the action is applicable, the observation is a statement about the
action being executed (e.g., ``I drive truck truck\_0 from ... to
...'' for the NL action ``Drive truck\_0 from ... to ...'').  This
description is obtained by converting the PDDL action into its NL
description using {\resourcename}.
%
%
If the action is not applicable, our simulator states why this is the
case, e.g., ``I cannot drive truck truck\_0 from location location\_0
in city city\_0 to ... because truck\_0 is not at location\_0.''
%
%
This type of feedback is constructed by converting the PDDL action and
the unsatisfied preconditions into their NL descriptions using the
template ``I cannot \#ACTION\# because \#UNSAT-PRE\#'' where each
\#PART\# is, again, constructed using {\resourcename}. For the creation of grammatical negations of unsatisfied preconditions, 
we determine where to add negations based on the position of the auxiliary verb (if there is one) or of the head of the phrase as found by a dependency parser \citep{qi2020stanza}.

Lastly, the simulator also determines whether the LLM's action choice
reached the goal which is used as a termination condition for LLM
policies.

\paragraph{\TLLM.} The \TLLM\ translates the NL actions from the \PLLM\
output to valid actions in the PDDL format so that it can be further
passed to the simulator. This translation is done by providing the
\TLLM\ a prompt consisting of a statement providing information about
the task (i.e., translating NL into PDDL) and the output format,
followed by the NL descriptions of all actions of the domain obtained
using {\resourcename} and the objects from the current problem.
Lastly, up to five pairs of an NL action snippet from the domain
and the corresponding PDDL action are included as few-shot examples.
The domain-specific prompts are generated entirely automatically based
on the generated PDDL-to-NL conversions and are identical for all
action choice mechanisms (see \appprompts\ 
for details).  This approach can be applied to any domain and is
independent of the order of the verb and its arguments in the NL
descriptions, hence allowing more flexibility than the domain-specific,
rule-based translation approach used by
\citet{valmeekam2023planbench}.

\begin{figure}[t]
    \centering
    \includegraphics[width=0.47\textwidth]{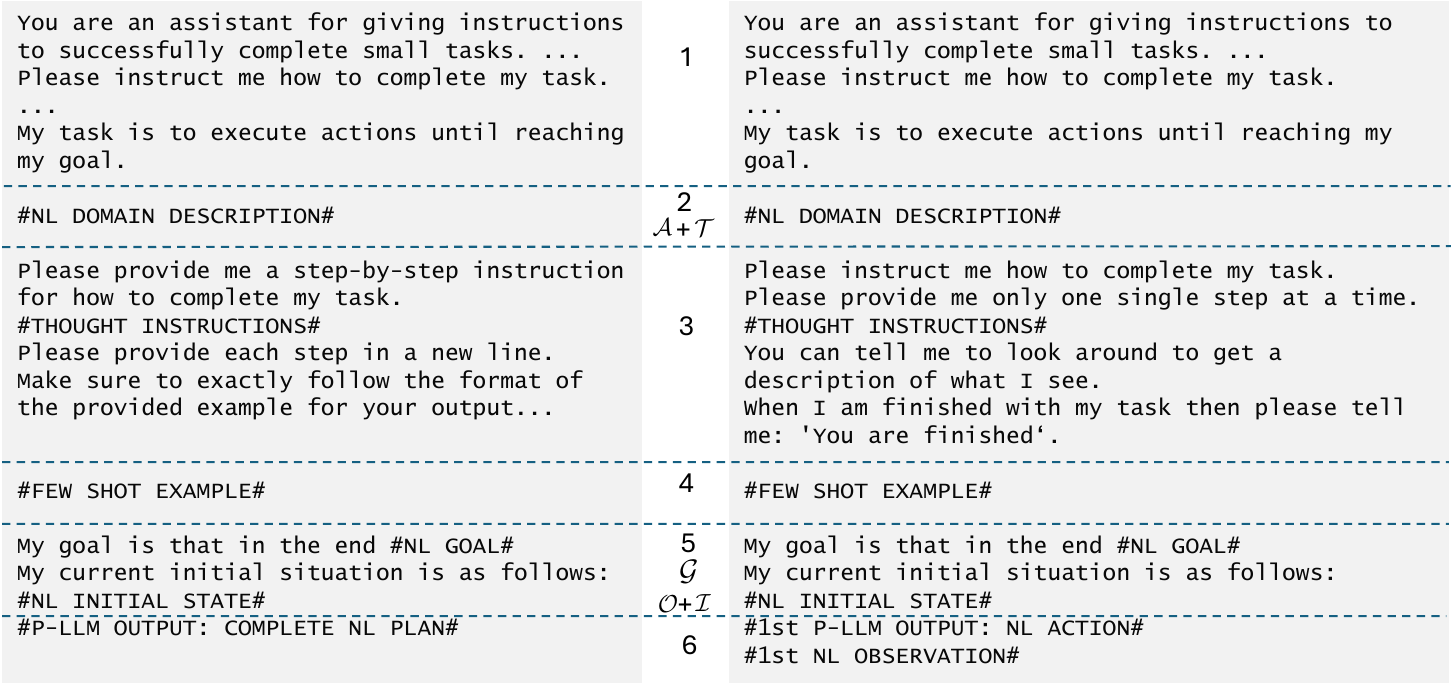}
    \caption{Structure of the prompts for the \PLLM\ in the LLM planning (left) set-up and in the policy set-up at the second prediction step (right). 
    }
    \label{fig:prompt_templates}
\end{figure}

\paragraph{\PLLM.} The \PLLM\ component implements the actual action choice
mechanism. 
Figure~\ref{fig:prompt_templates} shows the structure of the prompts for
the \PLLM\ in the LLM planning (left) and LLM policy (right) set-up.
All prompts start with an instruction of the planning task (1), followed by the NL description of 
the domain (2) and specific instructions for the output of the individual LLM action-choice mechanisms (3).
Then a few-shot example from the same domain is included (4).
It consists of the NL description of the goal of the example problem, the
initial state and an example for the generation of an action or action
sequence 
method and we discuss it in detail below.
%
At the end, the goal and initial state of the target
problem~(5) is described (see complete prompts in \appprompts).

%
The action-choice instructions (part (3)
and few-shot examples (4)
demonstrating the content and format expected from the \PLLM's output
are specific to the individual LLM action-choice mechanisms. We focus on 
four action-choice mechanisms:
two LLM planning techniques 
and two LLM policy techniques 
described
in what follows.

\subsection{LLM Planning}
For LLM planning, the \PLLM\ predicts a complete sequence of NL
actions, each in a separate line (see instructions (3)
in~Figure~\ref{fig:prompt_templates} (left)). The output of the
\PLLM\ is translated into PDDL line-by-line by the \TLLM\ and then
passed to the simulator that determines whether the generated action
sequence is a valid plan for the task. We include two LLM planning mechanisms: Basic and CoT. 

\begin{figure}[t]
    \centering
    \includegraphics[width=0.47\textwidth]{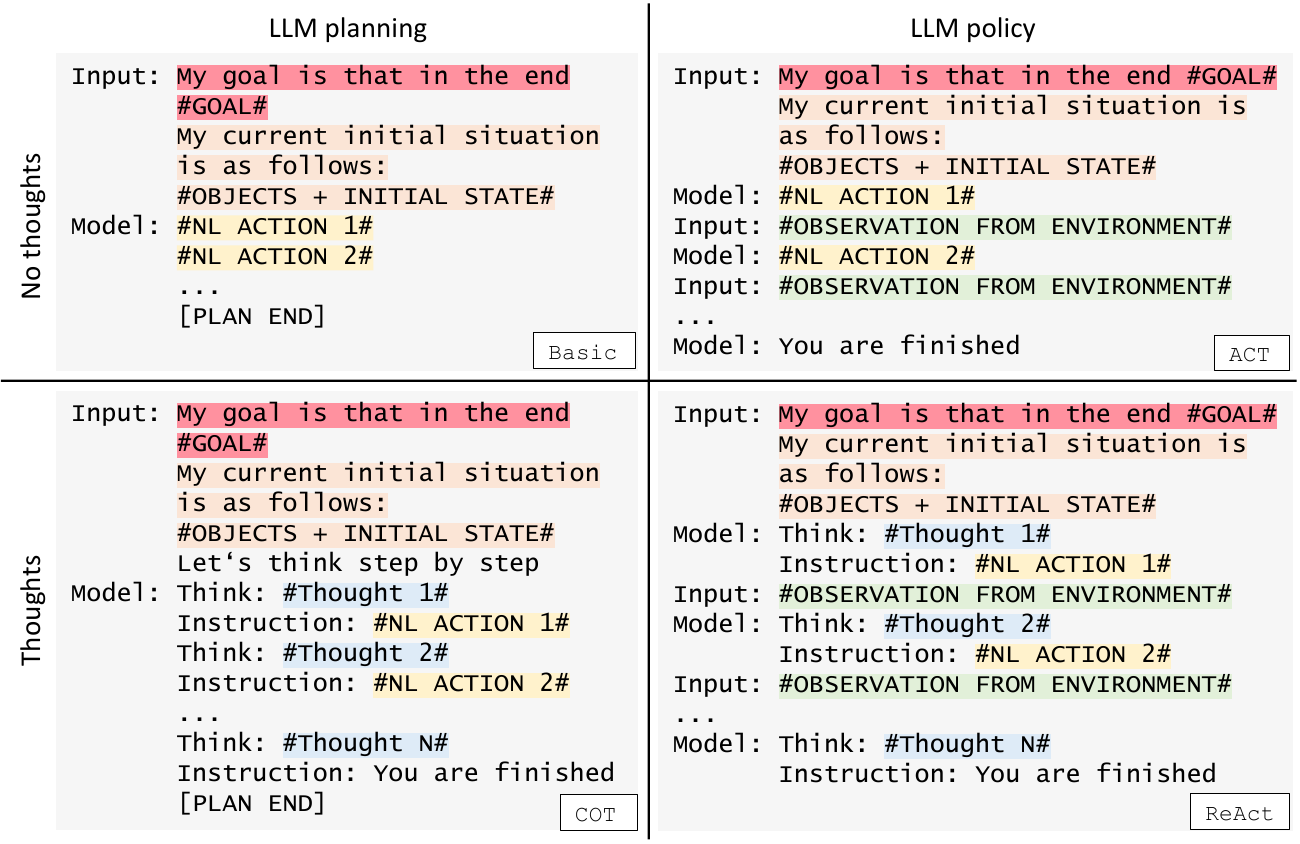}
    \caption{Structure of the few-shot examples for the four mechanisms. }
    \label{fig:few-shot_struct}
\end{figure}

\paragraph{Basic.} Here, we prompt the \PLLM\ to generate a complete plan of NL actions. We follow \citet{valmeekam2023planning} and present each action of the example plan in a separate line and include a special tag to signal the end of the plan as shown in Figure \ref{fig:few-shot_struct} (top left).

\paragraph{CoT.} \citet{NEURIPS2022_9d560961} showed that
prompting an LLM to generate a chain of thoughts, i.e., a sequence of
explicit reasoning steps, improves the results on a range of reasoning
tasks. The exact form and content of a thought are flexible and can include
explicit reasoning over the current state, the next action or a goal
to satisfy. For example, a thought in Logistics could be ``Now,
package\_0 is at truck\_1 and truck\_1 is at location\_2
in city\_4. Package\_0 needs to be moved to location\_3 in city\_4.''
The generation of thoughts by the \PLLM\ is elicited by adding thoughts
between the actions in the \Pex\ (see Figure \ref{fig:few-shot_struct},
bottom left) and additional instructions in the prompt (see part (3) in~Figure~\ref{fig:prompt_templates}).

\subsection{LLM Policies}
For LLM policies, the \PLLM\ generates a single action that is
directly translated by the \TLLM\ and passed to the simulator which in
turn produces a NL observation (or terminates the planning process in
case the goal is satisfied).  The generated NL observation is added to
the history buffer (see Figure \ref{fig:approach_overview}
(bottom)). In each step, the initial prompt for the \PLLM\ is extended
by all its previous outputs and observations from the history buffer
(see part (6) in~Figure~\ref{fig:prompt_templates} (right)).
This approach provides the LLM access to the history of the action choice
process as LLMs themselves do not include any memory, i.e., each call to an
LLM is independent.
Moreover, the process is not stopped when the \PLLM\ outputs an
inapplicable action as the observations provide information that
can be exploited by the LLM in subsequent steps.
We also equip the {\PLLM} with an option to ask for the current state which is
then provided by the simulator in the form of its NL description. 

We focus on two LLM policy techniques: ReAct and Act.

\paragraph{ReAct.} ReAct combines CoT reasoning with information received
from an environment.
It was originally proposed for interactive decision-making tasks \citep{yao2022react}.
At each step, the \PLLM\ predicts a thought and an action and receives an observation from the simulator (see Figure \ref{fig:few-shot_struct}, bottom right). 
The complete output of the \PLLM\ including the thought and the
observation are added to the history buffer. 

\paragraph{Act.} The Act mechanism works in the same way as ReAct but does not include reasoning thoughts (Figure \ref{fig:few-shot_struct}, top right). 

Note that all \Pex s are generated automatically by converting one small example
problem and its plan into NL. For the LLM policy mechanisms, the simulator is used to generate the corresponding observations. 
For CoT and ReAct, reasoning thoughts are required. 
We use an LLM-based approach to obtain them: first the \Pex\ in the ReAct structure is created (Figure \ref{fig:few-shot_struct}, bottom right), but with placeholders instead of actual thoughts.
We manually create thoughts for one example problem from Logistics and use this as few-shot example when prompting
an LLM to come up with appropriate thoughts to replace the placeholders for
other problems and domains. For the CoT \Pex s the observations get removed afterwards (see \appreact).

\section{Experiments}\label{sec:exp}

We evaluated all our action-choice mechanisms,
\exBasic (Basic), \exCoT, \exAct and \exReAct (ReAct), on different kinds
of input prompts, and against representatives of other planning mechanisms.
We used GPT-4o 
as LLM for all evaluated methods. 
The LLM policies \exAct and \exReAct are not guaranteed to terminate and
imposing a time limit is not an option here because we use GPT-4o via its API with
high variability of response times.
Therefore, we limit the number of steps for the LLM policies in each
instance by using twice the length of the plan
generated by satisficing greedy best-first search (GBFS) with the FF heuristic
\cite{hoffmann2001ff}.

All results for the LLM-based action choice methods are averaged over 5
runs; each run uses a different value of the ``seed'' parameter
of the GPT-4o API.
We compare our methods with PDDL2NL prompts against the following methods.

\paragraph{\exManual:}
We compare to the manually provided NL prompts designed by
\citet{valmeekam2023planning} for the domains Blocksworld, Depots and Logistics.

\paragraph{\exPDDL:}
Following the prior work of \citet{silver2022pddl},
we evaluate LLMs taking directly the PDDL input
instead of its NL descriptions.
We use only \exBasic and \exAct as it is not clear how to provide comparable
thoughts for PDDL inputs.
In this case, the {\TLLM} is skipped. Instead, the PDDL action is extracted
from the output of the {\PLLM} using a regular expression and is passed
directly to the simulator.
Additionally, the observations are modified to
``The action \#ACTION\# was successfully executed.'' and
``The action \#ACTION\# is not applicable because \#UNSAT-PRE\#.'' where
\#UNSAT-PRE\# is a concatenation of the unsatisfied PDDL predicates, each
followed by ``is true'' or ``is false''.
For a fairer comparison, we keep a slightly adapted task instruction as part of the
prompts and only replace the NL domain, goal and initial state descriptions
by their original PDDL input (see {\appprompts} for example prompts).
This is closer to the way in which
\citet{valmeekam2023planning} test PDDL inputs as they, in contrast to
\citet{silver2022pddl}, include the PDDL domain description and also a
short task description in natural language.

\paragraph{\exTemplate:}
As a baseline PDDL-to-NL translation, we use a simple template-based method
for converting PDDL into NL prompts:
Unary predicates ``(p ?x)'' are translated to snippets ``\{?x\} is `p''',
higher-arity predicates ``(p ?x1 ... ?xn)'' are translated to
``\{?x1\}, ..., and \{?xn\} are in relation `p''', and PDDL actions
``(act ?x1 ... ?xn)'' are
translated to ``apply the action `act' to object \{?x1\}, ..., and object
\{?xn\}'' where the word ``object'' is replaced with the parameter's type
when it is specified.
The composition of these snippets into the final prompt is done in the same
way as in \resourcename.
Since it is, again, not clear how to provide thoughts for this method, we
use it only with \exBasic and \exAct mechanisms.

\paragraph{\exRnd:}
To test whether LLM-based methods carry any information at all, we use a
simple random search (\exRnd) limited by the same number of steps as the
LLM policy that, in every step, selects an applicable action uniformly at
random. The results for \exRnd are averaged over 10 random seeds.

\paragraph{\exBFS, \exLMC, \exFF:}
We use breadth-first search (\exBFS) to get a sense of
hardness of tasks and how LLM-based methods compare to a trivial symbolic
baseline.
We also use two strong optimal and
satisficing planner baselines, namely A$^\star$ with the LM-cut heuristic
(\exLMC) \cite{lmcut}, and GBFS with the FF heuristic
(\exFF) \cite{hoffmann2001ff}.
These were run on Intel Xeon
E5-2687W processors with 30 minutes and 8 GB time and memory limits.

We use two benchmark sets.
First, we use 13 classical planning domains (see Table~\ref{tab:coverage}) including the three used by
\citet{valmeekam2023planning} and the Beluga domain which was published after the training of GPT-4o (\citeauthor{Eisenhut2024}, \citeyear{Eisenhut2024}). For Beluga, we select the 21 smallest problem instances from \citet{Eisenhut2024}.
In the other domains, we generate 21 relatively small
solvable instances with optimal plan lengths between 3 and 20.
We select a small instance (short plan; see details in
\appdom) for the P-LLM few-shot example (plan
or single action choice), and use the remaining 20 instances as our
benchmarks. 

Second, we also compare the best {\resourcename} action-choice mechanisms
to the symbolic baselines (\exRnd, \exBFS, \exLMC, \exFF) on a subset of
IPC benchmarks where it is possible and feasible.
First, we disregard domains with PDDL features unsupported by {\resourcename}
(e.g., conditional effects, quantifiers); and since LLM-based methods can rarely scale beyond \exFF (as we show later), we
remove instances that cannot be solved by \exFF within 30 minutes.
Apart from this, the main limiting factor is feasibility in terms of monetary cost for
GPT-4o calls. For 16 domains, the estimated cost (per domain!) was $> 150$USD; for those, we considered only the instances solved by BrFS, reducing the cost to $\leq 150$USD for 15 domains.
In addition to the monetary issues, we could not successfully run our
method on Agricola, Folding, Parking, Richochet-Robots, Sokoban, Tetris,
and Tidybot (e.g., in some cases PDDL2NL failed to generate NL snippets due to large numbers of action parameters).
Overall, the resulting IPC benchmark set consists of 675 instances in 41 domains.

\begin{table}[t]
\centering
{\def\arraystretch{.8}\tabcolsep=1.7pt
{\fontsize{8}{9}\selectfont
\begin{tabular}{l|rrrr|rr|rr||rrrr}
\multirow{2}{*}{Domains}
    & \multicolumn{8}{c||}{LLM-based Approaches} & & & & \\
    & \multicolumn{4}{c|}{\resourcename}
    & \multicolumn{2}{c|}{\exTemplate}
    & \multicolumn{2}{c||}{\exPDDL}
    & \multicolumn{4}{c}{Symb. Baselines}\\
    & \exBasic & \exCoT & \exAct & {\exReAct}
    & \exBasic & \exAct
    & \exBasic & \exAct
    & \exRnd & \exBFS & \exLMC & \exFF \\
\hline
Block.  & 13 & 15  & 17  & \textbf{18} & 8 & 12 & 13 & 14 & 1 & 20 & 20 & 20  \\
\hspace{.2em} $\rotatebox[origin=c]{180}{$\Lsh$}$\exManual & 15 & 14 & 17 & 18 & &  & & & & & &  \\
Logistics   & 5 & 10 & 16 & \textbf{19}  & 2 & 7 & 6 & 15 & 0 & 20 & 20 & 20\\
\hspace{0.2em} $\rotatebox[origin=c]{180}{$\Lsh$}$\exManual & 3 &5  & 17 & 12 &  & & & & & & & \\
Depot  & 0 & 5 & 5 & 13 & 0 & 5 & 0 & 3 & 0 & 20 & 20 & 20 \\
\hspace{0.2em} $\rotatebox[origin=c]{180}{$\Lsh$}$\exManual& 3 & 6 & 6 & \textbf{15} & & & & & & & & \\
Beluga & 4 & 4 & 8 & \textbf{9} & 3 & 5 & 5 & 3 & 0 & 20 & 0 & 20 \\
Ferry  & 7 & 12 & 14 & \textbf{18} & 0 & 13 & 8 & 17 & 0 & 20 & 20 & 20\\
Floortile  & 0 & 0 & 0 & 0 &0 & 0 & 0 & 0 & 0 & 18 & 20 & 20\\
Goldm.  & 1 & 1 & 3 & 1 & 1 & 3 & 1 & \textbf{4} & 0 & 20 & 20 & 20\\
Grid  & 8 & 6 & 16 & \textbf{18} & 1  &  12 &  6& 12 & 0 & 20 & 20&20\\
Grippers  & 9 & 17 & 17 & \textbf{20} & 10 & \textbf{20} & 12 & 19 & 0 & 20& 20& 20\\
Movie  & \textbf{20} & \textbf{20} & \textbf{20} & \textbf{20} & \textbf{20} &\textbf{20} & \textbf{20} & \textbf{20} & 3 & 20&20 &20\\
Rovers  & 0 & 0 & \textbf{18} & \textbf{18} & 1 & 17 & 1 & 11 & 1 & 20& 20& 20\\
Satellite  & 14 & 16 & \textbf{20} & \textbf{20} & 11 & 18 & 14 & 18 & 0 & 20& 20& 20\\
Visitall  & 19 & 19 & \textbf{20} & \textbf{20} &  18&\textbf{20}  & \textbf{20} & \textbf{20} & 8 & 20& 20& 20\\
\hline
$\Sigma$ (260) & 100 & 125 & 174 & \textbf{194} & 75 & 152 & 106 & 156 & 13 & 258 & 240  & 260\\
\hline
\hline
\multicolumn{12}{l}{Further scaled selected domains:} \\
\hline
Block. & 3 & 3 & 12  & \textbf{14} & 0 & 6 & 1 & 4 & 0 & 12 & 19 & 20 \\
Ferry & 0 & 0 & 7 & 15 & 0 & 9 & 0 & \textbf{17} & 0 & 8 & 13 & 20 \\
Gripper & 17 & 12 & \textbf{20} & 19 & 16 & \textbf{20} & 16 & \textbf{20} & 0 & 10 & 8 & 20 \\
Visitall & 9 & 2 & 16 & \textbf{18} & 7 & 17 & 14 & 16 & 1 & 10 & 18 & 20 \\
\hline
$\Sigma$ (80) & 29 & 17 & 55 & \textbf{66} & 23 & 52 & 31 & 57 & 1 & 40 & 58 & 80\\
\end{tabular}
}
}
\caption{\label{tab:coverage}%
Number of solved tasks out of 20 per domain. ``Valm23'' rows show
results of the manual encodings by \citet{valmeekam2023planbench}
in the respective domains.
We show in \textbf{bold} the best LLM-based method.
LLM-based results are averaged over 5 seeds, \exRnd over 10 seeds.}
\end{table}

\paragraph{Comparison to \exManual.}
Table~\ref{tab:coverage} shows the number of solved tasks (coverage) per
domain and overall.
The comparison in Blocksworld, Logistics and Depot to manual descriptions
(see \exManual rows)
shows that using our automatic translations results in
comparable performance.
The most significant difference can be observed for \exCoT and \exReAct
action-choice mechanisms in Logistics where our method solves 5 and 7 more
tasks, respectively.
This is a little bit surprising result considering that the hand-crafted
descriptions by \citeauthor{valmeekam2023planbench} contain additional
information that is not explicitly stated in PDDL
(e.g., that all locations within a city are directly connected)
which is thus not available in the NL descriptions automatically generated
by {\resourcename}.

\paragraph{Comparison between {\resourcename} variants.}
The results of \exBasic with the automatic translation over all
domains support previous findings that a basic prompting technique
does not work particularly well for plan generation
\cite[e.g.,][]{valmeekam2023planbench,liu2023llmp}. Adding reasoning
thoughts (\exCoT) improves performance substantially overall, though
the impact varies per domain and can also deteriorate performance (namely
in the Grid domain).

Using the LLM as an action policy instead of a plan generator in
\exAct yields a major performance boost, dominating \exBasic and
\exCoT consistently in every domain, with major coverage improvements
in many domains. This shows that the use of LLMs, not for plan
generation, but as \emph{a part of plan generation} works much better---in
this case, the LLM being used for action choice only,
with the computation of states being done symbolically and fed back
into the LLM prompts.
Adding reasoning thoughts to \exAct in \exReAct yields another
performance boost, consistently dominating coverage across all four
LLM action choice mechanisms (with an exception in Goldminer) and achieving
best LLM performance.


\paragraph{Comparison to \exPDDL and \exTemplate.}
\exPDDL performs slightly better than PDDL2NL for the \exBasic variant,
whereas it is the other way around for \exAct.
Nevertheless, \exReAct with PDDL2NL is clearly superior mainly because it
uses natural sounding, intuitive thoughts in NL.
It is not clear how we could obtain such thoughts for \exPDDL.

The simple template-based method (\exTemplate) works decently in some domains
(e.g., Depot, Goldminer, or Visitall), but PDDL2NL generates NL task
descriptions that are at least as good (and often better) in most
domains.
We think this is because LLMs ``understand'' language and therefore
the NL descriptions generated by PDDL2NL are more naturally-sounding than
the \exTemplate descriptions.
In comparison to \exPDDL, \exTemplate is clearly the weaker method.
It also seems that \exTemplate works well in domains where also \exPDDL
works well, and whenever PDDL2NL works significantly better than
\exTemplate, it also works better than \exPDDL.

We additionally ran experiments on a version of Blocksworld where the structure of the domain is preserved but the actions and predicates have only very generic names.
In particular, we replaced all predicate names with \mbox{``predicate1''}, ...,
``predicate$N$'' and action names with \mbox{``action1''}, ..., ``action$M$'', where
$N$ and $M$ are the number of predicates and actions in the PDDL file, respectively.
This resulted in very generic, template-like, NL snippets produced by the LLM (e.g. ``predicate3 is true'', ``perform action1 on an object \{?obj\}''). Running the \exBasic and \exAct \mbox{approaches} on the generated prompts yielded 0 coverage for both \mbox{approaches}. 

\paragraph{Comparison to symbolic baselines.}
The comparison to the random baseline \exRnd clearly shows that LLM methods
are able to extract at least some useful information from the task
descriptions (with the exception of Floortile where all LLM methods failed).
%
As can be seen from the \exBFS results, the evaluated tasks are fairly
small, and yet LLM methods fail to solve them all.
The performance of LLM-based methods significantly lag behind symbolic
methods: \exFF solves all tasks, \exLMC solves all tasks except for the \mbox{Beluga} domain, and even
\exBFS solves all tasks except for two in Floortile
(the average runtime of \exLMC, \exFF and \exBFS was
$0.2$, $0.1$ and $6$ seconds, respectively). 
This behaviour was already observed before---here, we provide a more
comprehensive evaluation enabled by the automatic generation of NL
descriptions.
Nevertheless, we can also see some encouraging results with \exReAct in
many domains.

\paragraph{Scaling experiments.}
To see how far the LLM action choice mechanisms can scale,
we conduct more experiments with larger generated tasks.
We focus on Blocksworld, Ferry, Grippers, and Visitall because \exReAct
performs very well there and it is easy to scale these domains with a
single parameter.

The scaled data for these domains is created as follows.
For each domain, we randomly generate a set of tasks, always varying only a
single parameter: the number of blocks, cars, balls and locations for
Blocksworld, Ferry, Grippers and Visitall, respectively (see \appdom\ for
more details).
Then we run \exBFS and \exLMC on each task with 30 minutes and 8 GB limits.
Then we identify the value $N$ of the varied parameter (e.g.,
number of balls in Grippers) at which either \exBFS or \exLMC is unable to
solve the task.
For the final dataset we select 20 problems per domain for which the scaled
parameter values are around the identified threshold $N$.
We use the same few-shot examples as in the first round of experiments. 
%
%
The bottom part of Table~\ref{tab:coverage} shows the coverage on the scaled
benchmark set.  

\exRnd can solve only a single task in Visitall, confirming
the vast superiority of LLM-based methods as much more informed.
\exBFS\ is much better, but is also challenged by the size of these
tasks. It is outperformed by \exReAct\ in all domains.
While \exBFS\ is a
very basic symbolic baseline, this provides additional evidence of
\exReAct's planning abilities.
Indeed, remarkably, \exReAct outperforms the state-of-the-art
optimal planner \exLMC in 2 domains (and matches its performance in 1).
This is remarkable given that \exReAct,
in contrast to \exLMC, does not perform any search. On the other hand,
it should be noted that Ferry and Grippers are structurally simple
domains, and that \exReAct---in contrast to \exLMC---does not give any
plan-optimality guarantee. 
The satisficing planner \exFF, which is comparable to \exReAct\ in
that regard, still has perfect coverage also on these scaled tasks, so
the benchmarks are still not ``hard enough'' to be challenging for
satisficing planning.

Overall, while LLM action choice at this point lags far behind
symbolic planners, there are isolated islands of good performance, and
our results do show promise for LLM planning abilities, in particular
if used as part of a larger symbolic planning machinery (as in case of
\exAct and \exReAct).

\begin{table}[t]
\centering
{\def\arraystretch{.9}\tabcolsep=2.3pt
{\fontsize{8}{9}\selectfont
\begin{tabular}{l|rr|rrrr}
\multirow{2}{*}{Domains}
    & \multicolumn{2}{c|}{\resourcename}
    & \multicolumn{4}{c}{Symbolic Baselines}\\
    & \exCoT & {\exReAct}
    & \exRnd & \exBFS & \exLMC & \exFF \\
\hline
barman11/14 (10)  & 0 & 3 & 0 & 10 & 3 &10 \\
blocks00 (35) & 3 & 22 & 0 & 21 & 28 & 35 \\
childsnack14 (16)  & 6  & 15 & 0 & 0 & 0 & 16\\
gripper98 (19)  & 12  & 19 & 0 & 7& 6& 19\\
logistics98/00 (29)  & 1  & 28 & 0& 12& 21& 29\\
movie98 (29) & 29 & 29 & 0 & 29 & 29 &  29 \\
rovers06 (6)  & 1  & 5 & 0 &6 &6 &6 \\
satellite02 (5)  & 1  & 4 & 0 &5 &5 &5 \\
transport08/11 (31)  & 3  & 23 & 0 & 18&19 &31 \\
visitall11/14 (13)  & 6  & 13&0 &13 &13 &13 \\
\hline
others (482 in 27 domains) & 4 & 18 & 1 & 291 & 311 & 482 \\
\hline
$\Sigma$ (675) & 66 & 179 & 1 & 412 & 441 & 675 \\
\end{tabular}
}
}
\caption{\label{tab:coverage-ipc}%
Number of solved tasks for selected IPC instance suites.
``others'': sum over IPC instance suites where both \exCoT and \exReAct solved less than
25\% of instances.}
\end{table}

\paragraph{IPC benchmarks.}
We also conducted experiments on a subset of IPC domains with the best
plan-generation (\exCoT) and policy (\exReAct) methods (see
Table~\ref{tab:coverage-ipc}).
\exCoT and \exReAct solved less than 25\% in 32 and 27 IPC instance suites,
respectively, and 0 tasks in 29 and 20 instance suites, respectively.
\exCoT solved all tasks solved by \exFF only in movie98, but \exReAct
matched the performance of \exFF in movie98 and gripper98
(and 3 more suites limited to instances solved by \exBFS), and solved only
one less task in childsnack14 and logistics00 (and 2 more suites limited
by \exBFS).
This shows that LLM-based methods are
usually unable to scale to larger instances, but also that there are
domains where \exReAct 
is able to achieve a good performance.
In fact, it seems that \exReAct either works fairly well or not at all (with
very few outliers).

\paragraph{Cost analysis.} We additionally conduct a comparison of the cost for the different input prompt versions in terms of the total number of input tokens processed by the \PLLM\ on solved problem instances. In particular, we compute the average number of input tokens for each domain in Table~\ref{tab:coverage}, considering only those problem instances that were solved by \exAct for each of \resourcename, \exTemplate and \exPDDL.
The sum of the per-domain averages for inputs generated by \resourcename\
is 816~732 tokens and hence cheaper than the \exTemplate\ approach with
1~064~573 tokens. In comparison to PDDL inputs, our approach is more costly
(816~732 vs. 665~881), but by only up to a factor of 1.5 per domain. 

We also compare our automatic approach with the human-generated input
prompts from \citet{valmeekam2023planbench}. The sums of the average number
of tokens for commonly solved instances of Act for \resourcename\ and
\exManual\ are 69~137 vs. 52~080, respectively. Hence, our automatic approach is more costly but only slightly and only up to a factor of 1.5 per domain. We provide the per-domain averages in the extended version of this paper \appcost.

\section{Conclusion}
\label{sec:conclusion}

LLMs are rapidly gaining prominence in many sub-areas of AI, and the
question if and how they can be applied in AI Planning is highly
relevant. Following up on previous work in this direction, we show how
to automate the conversion of PDDL into natural language prompts.
Based on this, we contribute broad experiments, highlighting that the
automatic conversion does not result in a performance loss relative to
the previous hand-crafted prompts, and examining performance relative
to representative symbolic methods. The results enhance our knowledge
of LLM action choice performance, and demonstrate convincingly that
LLMs do have \emph{some} action-choice ability, outperforming random
action selection and, in a few cases, even a state-of-the-art optimal
planner. This performance is still far from the state of the art in
symbolic (satisficing) planning, yet it is achieved without any
search, pointing to the potential of more general uses of LLMs in
planning.

The most direct question for future work, in our view, is how to
combine LLMs with symbolic search methods. Our work lays the basis for
that through the automatic translation of PDDL into natural language
prompts, which as our results show boosts the LLM's planning ability.
The space of possible combinations is vast. One could use the LLM to
suggest preferred actions in search, one could search around
LLM-predicted plans or actions, one could apply plan repair to the
LLM's suggestion (as suggested by \cite{valmeekam2023planning} with
LPG \cite{gerevini2003planning}), one could use LLM-generated plans as
the basis for heuristic functions, etc. For further research on the
question whether LLMs on their own (without search) can yield better
planning performance, specialized training or neurosymbolic methods
may be interesting to look at.

\section{Acknowledgments}
The work was partially funded by the Deutsche Forschungsgemeinschaft (DFG, German Research Foundation) under the project number 232722074 – SFB 1102. We gratefully acknowledge the stimulating research environment of the GRK 2853/1 “Neuroexplicit Models of Language, Vision, and Action”, funded by the Deutsche Forschungsgemeinschaft under project number 471607914.

\bibliography{mybibfile}

\appendix
\input{appendix/1_experiments}
\input{appendix/2_cost_analysis}

\input{appendix/3_autoplanbench_prompts}

\input{appendix/4_planning_prompts}

\input{appendix/5_react_few_shot}

\end{document}

%% file: appendix/1_experiments.tex
\clearpage
\section{Experimental set-up: LLM models}\label{app:exp}

All experiments with GPT-4o were conducted using the OpenAI API and the gpt-4o-2024-08-06 model. For the \exBasic and \exCoT approach we use the OpenAI Batch API. All experiments were conducted in October 2024, except for the experiments on the Beluga domain which were conducted in Februrary 2025.  

For all models and tasks we set the temperature to 0.0 and use caching to reduce the cost: If the model has already received the same input (including the prompt, the user message and the history) before, the previously generated response text is retrieved from the cache. This affects in the first place the \TLLM\ because all planning approaches and individual problems have different initial prompts and therefore the P-LLM should never get exactly the same input twice. 
We set the seed parameter to values between 1 and 5. 
For all other parameters we keep the default values except for the maximum number of tokens:
\begin{itemize}
    \item \APBLLM: 200 
    \item \PLLM: no limitation
    \item \TLLM: 750
    \item LLM to generate thoughts for planning example: 2000
\end{itemize}

All few-shot examples for the \APBLLM, \PLLM, {\TLLM} and the thought generation were not included in the system prompt but as part of the model history because we found that this improves the compliance with the required output formatting. The ``Input:'' and ``Model:'' prefixes in the presented prompts are used to indicate which parts were added as user prompts (``Input:'') and assistant prompts (``Model:'') respectively. Everything occurring before the few-shot examples is provided as part of the system prompt.

\section{PDDL domains and problems}\label{app:domain}

For our experiments we select 13 PDDL domains from classical planning. We only select STRIPS domains with only ``not'' and ``and'' as operators and we removed the costs if the domain includes action costs. For the Blocksworld, Logistics and Depot domain we use the domain PDDL files and problems from the PlanBench repository\footnote{\url{https://github.com/karthikv792/LLMs-Planning/tree/main/plan-bench}}. For the Beluga domain, we select the 21 smallest problem instances from \citet{Eisenhut2024}. For all other domains we take the domain files from the PDDL-Generators repository\footnote{\url{https://github.com/AI-Planning/pddl-generators/tree/main}} \cite{seipp-et-al-zenodo2022} which offers a collection of planning domains and the problem generators for the domains. We use the available problem generators to generate the problems included in our dataset and select rather small values for the parameters that can be varied for the individual domains for creating the dataset for the main experiment.

The Movie domain is the only domain to which we make changes beyond removing costs. In particular, we modify the domain such that the actions for buying snacks have additional preconditions that do not allow buying them in all states. Additionally, we introduce some variation to the planning task by generating not only problems where one object of each kind of snacks needs to be bought but also problems were less (but at least one) kinds of snacks must be bought.


\paragraph{Scaling experiments.} For the scaling experiment, we focus on
scaling only one specific type of object per domain. In the Blocksworld
domain, the only parameter that can be scaled is the number of available
blocks. For Grippers, we only vary the number of balls while fixing the
number of robots to 1 and the number of rooms to 2, hence essentially
restricting the problems to problems of the Gripper domain. For the Ferry
domain, we fix the number of locations to 3 while varying the number of
available cars. Finally, for the Visitall problems we vary the $x$ and $y$
dimension of the grid and restrict the problems to those where exactly the
$x\times y$ locations need to be visited. For the Blocksworld, Gripper and
Visitall domains, always all objects of the varied type need to be part of a successful plan. For the Ferry domain, we select only problems where more than 80\% of the cars need to be moved to achieve the goal. 

\paragraph{IPC Domains.} We remove the action costs from all domains that do not have uniform action costs. Table \ref{tab:coverage-scaled-app} shows the full set of results on the IPC domains included in our experiments. Results in the upper part of the table were obtained on all instances that were solved by \exFF\ and results in the lower part were obtained only on those instances that were solved by \exBFS.

\begin{table}[ht]
\centering
\caption{Number of solved tasks for selected IPC domains. The upper part shows the results on domains where all tasks solved by \exFF\ were considered, the lower part the domains for which only tasks solved by \exBFS\ were included.}
{\def\arraystretch{.9}\tabcolsep=2.3pt
{\fontsize{8}{9}\selectfont
\begin{tabular}{lr||rr||rrrr}
\multirow{2}{*}{Domains} &
    & \multicolumn{2}{c|}{\resourcename}
    & \multicolumn{4}{c}{Symbolic Baselines}\\
    &&  \exCoT & {\exReAct}
    & \exRnd & \exBFS & \exLMC & \exFF \\
\hline
 blocks00 &(35)  & 3  & 22 & 0 & 21 & 28 & 35 \\
 childsnack14 &(16)  & 6  & 15 & 0 & 0 & 0 & 16\\
data-network18 &(19)  &  0    & 0 & 0 & 13 & 19 & 19  \\
depot02 &(17)  & 0  & 3 & 0 & 5& 6& 17\\
 elevators08 &(29)  & 0  &0 & 0 & 13 & 19 & 29 \\
elevators11 &(20)   & 0    & 0 & 0 & 12 & 17 & 20   \\
floortile11 &(9)  & 0  & 0 & 0 & 2 & 6 & 9 \\
floortile14 &(9)  & 0  & 0 & 0 & 0 & 5 & 9 \\
ged14 &(19)   & 3 & 3 & 0 & 19 & 19 & 19 \\
 gripper98 &(19)  & 12  & 19 & 0 & 7& 6& 19\\
hiking14 &(19) & 0 & 0 & 0 & 12 & 8 & 19  \\
labyrinth23 &(5)  &0 &0 & 0 & 5 & 1 & 5 \\
 logistics00 &(27)  & 0  & 26 & 0& 10& 19& 27\\
movie98 &(29)  & 29 & 29 & 0 & 29 & 29 &  29 \\
mprime98 &(34)  & 0 & 0 & 0 & 20 & 22 & 34  \\
mystery98 &(17) & 0 & 0 & 0 & 14 & 15 & 17  \\
 parking11 &(19)  & 0  & 0 & 0& 0& 2 & 19 \\
 pegsol08 &(29)  & 0  &0 & 0 & 27 & 26 & 29 \\
 pegsol11& (20)  & 0  & 0 & 0 & 18 & 17 & 20 \\
  scanalyzer08 &(27)  & 0  & 2 & 0 & 11 & 9 & 27 \\
scanalyzer11 &(17) & 0 & 0 & 0 & 9 & 7 & 17  \\
transport11 &(20)  & 0  & 15 & 0 & 7 & 8 & 20  \\
 transport14 &(20)  & 0  & 4 & 0 & 7& 7 & 20\\
  woodworking08 &(29)  & 0  & 1& 0 & 8 & 17 & 29\\
woodworking11 &(20)  &0& 0 & 0 & 4 & 13 & 20  \\
zenotravel02 &(19) & 1 & 4 & 0 & 7 & 12 & 19  \\
\hline\hline
barman11& (7)  & 0  & 2 & 0 & 7 & 3 &7 \\
barman14& (3)  & 0  & 1 & 0 & 3 & 0 & 3 \\
driverlog02& (7)  & 0  &  1 & 0 & 7 &7 &7 \\
freecell00& (19)  & 0  & 0 & 1 & 19&14 &19 \\
logistics98& (2)  & 1  & 2 & 0 & 2 & 2& 2\\
pipesw.-notank.04& (18) & 0 & 0&0 &18 & 17&18 \\
pipesw.tank.04& (13)  & 0 &0 & 0 & 13 & 10 & 13 \\
rovers06 &(6)  & 1  & 5 & 0 &6 &6 &6 \\
satellite02& (5) & 1  & 4 & 0 &5 &5 &5 \\
snake18 &(12)  & 0 & 0 & 0 &12 &6 &12 \\
termes18 &(11)  & 0  & 0 &0& 11 & 5 &11 \\
tpp06 &(5)  & 0  & 0 & 0 & 5 &5 &5 \\
transport08 &(11)  & 3  & 8 & 0 & 11&11 &11 \\
visitall11& (9)  & 5  & 9&0 &9 &9 &9 \\
visitall14 &(4)  & 1 & 4 & 0 & 4& 4& 4\\
\hline
$\Sigma$  & 675 &  66  & 179 & 1 & 412 & 441 & 675 \\
\end{tabular}
}
}
\label{tab:coverage-scaled-app}
\end{table}

\paragraph{Few-shot example selection.} For all experiments, we select a
small instance from the same domain as the few-shot example if available.
In particular, we select an instance where the plan generated by \exFF or
\exLMC has a length of five. If there is no such instance available in the
considered benchmarks then we select the instance for which the length of
the generated plan is closest to five. 

\begin{table*}[ht]
    \centering
    \small
    \caption{Natural language fragments generated by \resourcename\ for the predicates and examples of the Logistics domain.}
    \begin{tabular}{|l|l|}
    \hline
        \textbf{PPDL} & \textbf{NL description}  \\
        \hline
         (OBJ ?obj) & \{?obj\} is an object\\ 
         (TRUCK ?truck) & \{?truck\} is a truck \\ 
         (LOCATION ?loc) & \{?loc\} is a location \\ 
         (AIRPLANE ?airplane) & \{?airplane\} is an airplane\\ 
         (AIRPORT ?airport) & \{?airport\} is an airport\\ 
         (CITY ?city) & \{?city\} is a city\\ 
         (at ?obj ?loc) & \{?obj\} is at \{?loc\}\\ 
         (in ?obj1 ?obj2) & \{?obj1\} is in \{?obj2\}\\ 
         (in-city ?obj ?city) & \{?obj\} is in the city \{?city\}\\ 
         \hdashline
         (load-truck ?obj ?truck ?loc) & load object \{?obj\} into truck \{?truck\} \\ 
         & at location \{?loc\}\\
         (load-airplane ?obj ?airplane ?loc) & load object \{?obj\} into airplane \{?airplane\} \\
         & at location \{?loc\}\\ 
         (unload-truck ?obj ?truck ?loc) & unload object \{?obj\} from truck {?truck} \\
         & at location \{?loc\}\\ 
         (unload-airplane ?obj ?airplane ?loc) & unload object \{?obj\} from airplane \{?airplane\} \\
         & at location \{?loc\}\\ 
         (drive-truck ?truck ?loc-from ?loc-to ?city) & drive truck \{?truck\} from location \{?loc-from\} in city \\
         & \{?city\} to location \{?loc-to\} in the same city\\ 
         (fly-airplane ?airplane ?loc-from ?loc-to) & fly airplane \{?airplane\} from airport \{?loc-from\} \\
         & to airport \{?loc-to\}\\ 
         \hline
    \end{tabular}
    
    \label{tab:templates_example}
\end{table*}

\section{Conversion prompts and examples}\label{app:apb}

Figure \ref{fig:apb_prompt_pred} and \ref{fig:apb_prompt_act} show the prompts used to derive the core natural language snippets for the predicates and the actions based on which the overall domain and problem descriptions and the few-shot examples for the \PLLM\ are generated. For the \APBLLM\, i.e. for the conversion, we use the same few-shot examples for each input domain. The few-shot examples for both the predicate and the action conversion were created manually. 
The last few-shot example for the action ``move'' in Figure \ref{fig:apb_prompt_act} was designed specifically in order to avoid that the LLM picks up the pattern that the order of parameters in the natural language snippets needs to match the order of parameters in PDDL. 

The last part of Figure \ref{fig:apb_prompt_pred} shows the feedback that the \APBLLM\ receives when making a formal mistake in its prediction, i.e. if the NL snippet does not contain exactly the same variables as the PDDL input, if question marks are missing or the brackets are not set correctly. The same feedback message is used for correcting action snippet predictions.

Table \ref{tab:templates_example} presents the NL snippets generated by \resourcename\ for all PDDL predicates and actions of the Logistics domain as an example of the output of our conversion approach. 

\section{\PLLM\ and \TLLM\ prompts}\label{app:planning_prompts}

Figure \ref{fig:P-LLM_prompt_basic}, \ref{fig:P-LLM_prompt_cot}, \ref{fig:P-LLM_prompt_act} and \ref{fig:P-LLM_prompt_react} present the structures of the prompts for the \PLLM\ in the four LLM planning and LLM policy set-ups.  
When using PDDL as input instead of NL, we use slightly adapted versions of the prompts for the \PLLM\, as presented in Figure \ref{fig:prompt_basicpddl_struct} and \ref{fig:prompt_actpddl_struct}. Figure \ref{fig:prompt_ex_basic}, \ref{fig:prompt_ex_react} and \ref{fig:prompt_ex_pddl_act} show examples of the complete input prompts for the \PLLM\ in the Basic and ReAct set-up with NL and Act with PDDL as input. 

The prompts for the \TLLM\ are automatically created by \resourcename\ following the structure in Figure \ref{fig:t_llm_prompt_template}. For each domain, the pairs of PDDL and NL encodings of all actions of the domain are included in the prompt. Additionally, \resourcename creates few-shot examples for the translation from NL to PDDL by randomly sampling up to five distinct actions with different numbers of parameters if possible and sampling the required number of different objects from a list of example object names. The example object names were designed such that they are similar to object names from the actual planning problems but unlikely to overlap with them. The example objects are included in the list of available objects in the prompt which is extended by the actual objects of the specific problem instance. Except for the objects list, the prompt is the identical for all problem instances of the same domain. In particular, there are no differences between the different action choice mechanisms with respect to the prompt for the \TLLM.
Figure \ref{fig:t_llm_prompt} shows an example of the prompt for problems in the Blocksworld domain. 

\section{Automatic thought generation}\label{app:though_gen}

The few-shot examples for the \texttt{ReAct} and \texttt{CoT} approach require the availability of appropriate reasoning thoughts for the steps in the example problem. In order to automate the few-shot example generation, we let GPT-4o generate these thoughts in advance. Figure \ref{fig:thought_gen_temp} shows the structure of the prompt that is used for the thoughts generation: first, the instructions for the task are given, then a few-shot example for the task of generating thoughts is provided. The few-shot example is always the same example from the Logistics domain and consists of the NL domain description, the goal description, the actions of the optimal plan, the observations from the simulator and placeholders for the thoughts. The actual thoughts are included below and were manually created by us. 
Figure \ref{fig:thought_gen1} shows an excerpt of the actual prompt for generating thoughts for an example problem from Blocksworld based on the manually created Logistics example. 

%% file: appendix/2_cost_analysis.tex
\section{Cost analysis}

We conduct a comparison of the cost for the different input prompt versions
in terms of the number of input tokens processed by the {\PLLM}. For each of the 13 domains from our primary experiments, we select all problem instances that were solved by \exAct with all three different input versions, i.e. the automatically generated NL prompts, the templates and when using PDDL directly. We then compute the average number of input tokens per problem instance for each domain. Note that for LLM policies the input prompt at each action prediction step includes all input and output tokens from all previous prediction steps. 

Table~\ref{tab:cost_main} presents the per-domain averages when using the automatically generated input prompts (\resourcename), the templates (\exTemplate) and the PDDL input prompts. 

Table~\ref{tab:cost_valm} shows the per-domain averages for input prompts generated by our automatic approach (\resourcename) and on the human-generated inputs from \citet{valmeekam2023planbench} when considering all problem instances solved by \exAct\ with both input versions.

\begin{table}[ht]
    \centering
        \caption{Average number of input tokens processed by the \PLLM\ for a problem instance when running the \exAct\ approach on input prompts generated by out automatic approach (\resourcename), by templates (\exTemplate) and on PDDL inputs directly.}
    \begin{tabular}{l|r|r|r}
        Domain & \resourcename & \exTemplate  & \exPDDL \\
    \hline
       Beluga       & 68,308    & 102,909 & 59,546 \\
       Blocks.      & 7,948     & 11,344 & 6,354 \\
       Depot        & 27,144    & 41,497 & 23,732 \\
       Ferry        & 18,027    & 29,356 & 14,210 \\
       Floortile    & ---       & ---    & --- \\
       Grid         & 20,194    & 28,911 & 22,083 \\
       Grippers     & 23,791    & 29,002 & 16,331 \\
       Goldminer    & 17,591    & 20,962 & 25,731 \\
       Logistics    & 13,920    & 21,148 & 12,060 \\
       Movie        & 10,409    & 10,852 & 8,020 \\
       Rovers       & 54,646    & 67,693 & 42,353 \\
       Satellite    & 15,410    & 20,469 & 14,422 \\
       Visitall     & 6,490     & 7,430 & 6,541 \\
       \hline
        \hline
       \multicolumn{4}{l}{Further scaled selected domains:} \\
       \hline
       Blocks.      & 17,349    & 25,775 & 12,844 \\
       Ferry        & 181,908   & 259,394 & 145,710 \\
       Grippers     & 191,057   & 219,342 & 138,804 \\
       Visitall     & 142,541   & 168,491 & 117,140 \\
       \hline
       $\Sigma$     & 816,732   & 1,064,573 &  665,881\\
    \end{tabular}
    \label{tab:cost_main}
\end{table}

\begin{table}[ht]
    \centering
    \caption{Average number of input tokens processed by the \PLLM\ for a problem instance when running the \exAct\ approach on input prompts generated by out automatic approach (\resourcename) and on the human-generated inputs from \citet{valmeekam2023planbench}.}
    \begin{tabular}{l||r|r}
       Domain  & \resourcename & Valmeekam \\
       \hline
       Blocksworld & 9,291 & 8,445\\
       Depot & 35,352 & 24,346 \\
       Logistics & 24,494 & 19,289 \\
       \hline
       $\Sigma$ & 69,137 & 52,080 \\
    \end{tabular}
    
    \label{tab:cost_valm}
\end{table}

%% file: appendix/3_autoplanbench_prompts.tex
\begin{figure*}[ht]
    \centering
    \includegraphics[width=0.75\textwidth]{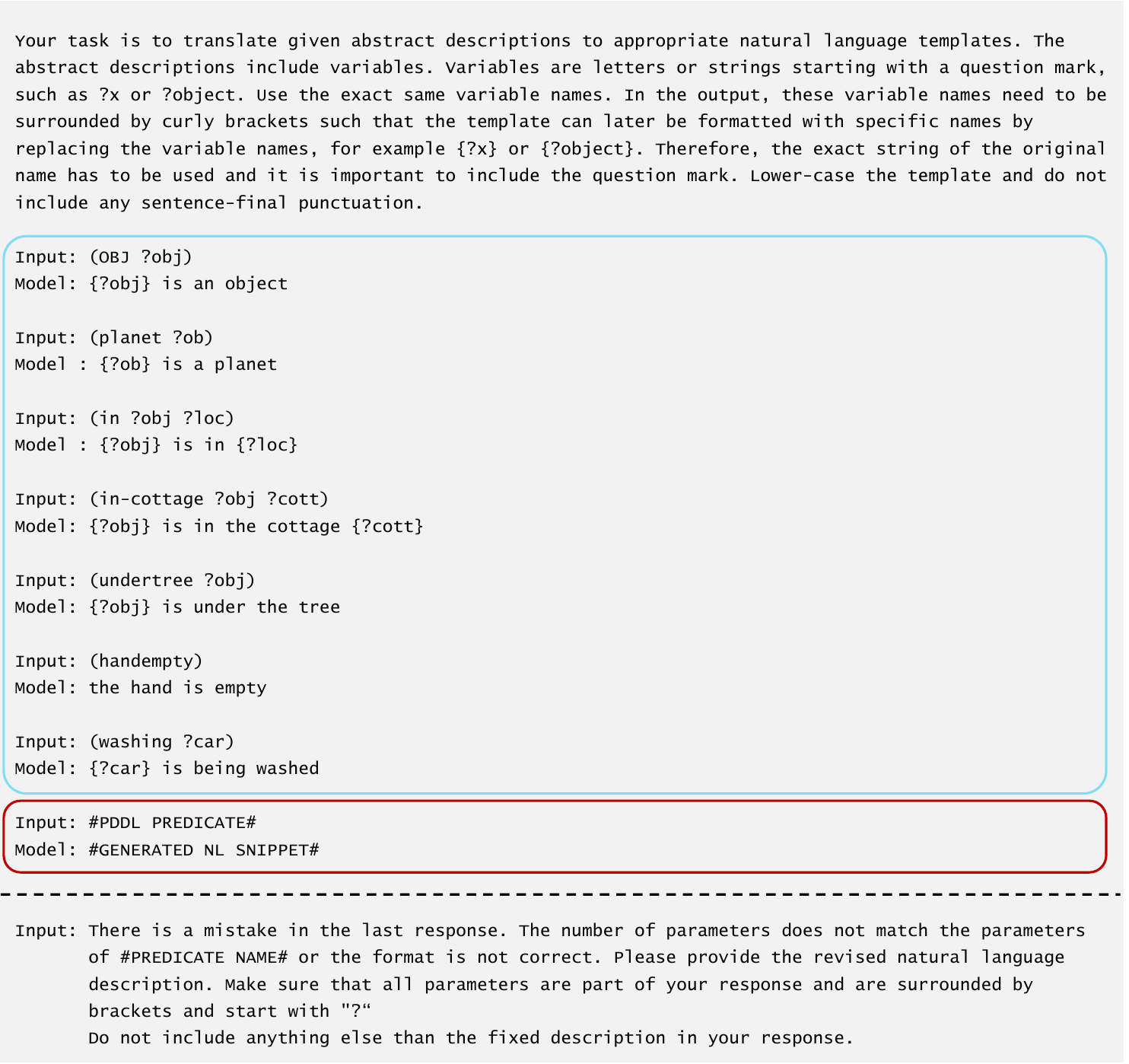}
    \caption{The complete prompt for the \APBLLM, used to generate the NL snippets for the PDDL predicates.}
    \label{fig:apb_prompt_pred}
\end{figure*}

\begin{figure*}[ht]
    \centering
    \includegraphics[width=0.75\textwidth]{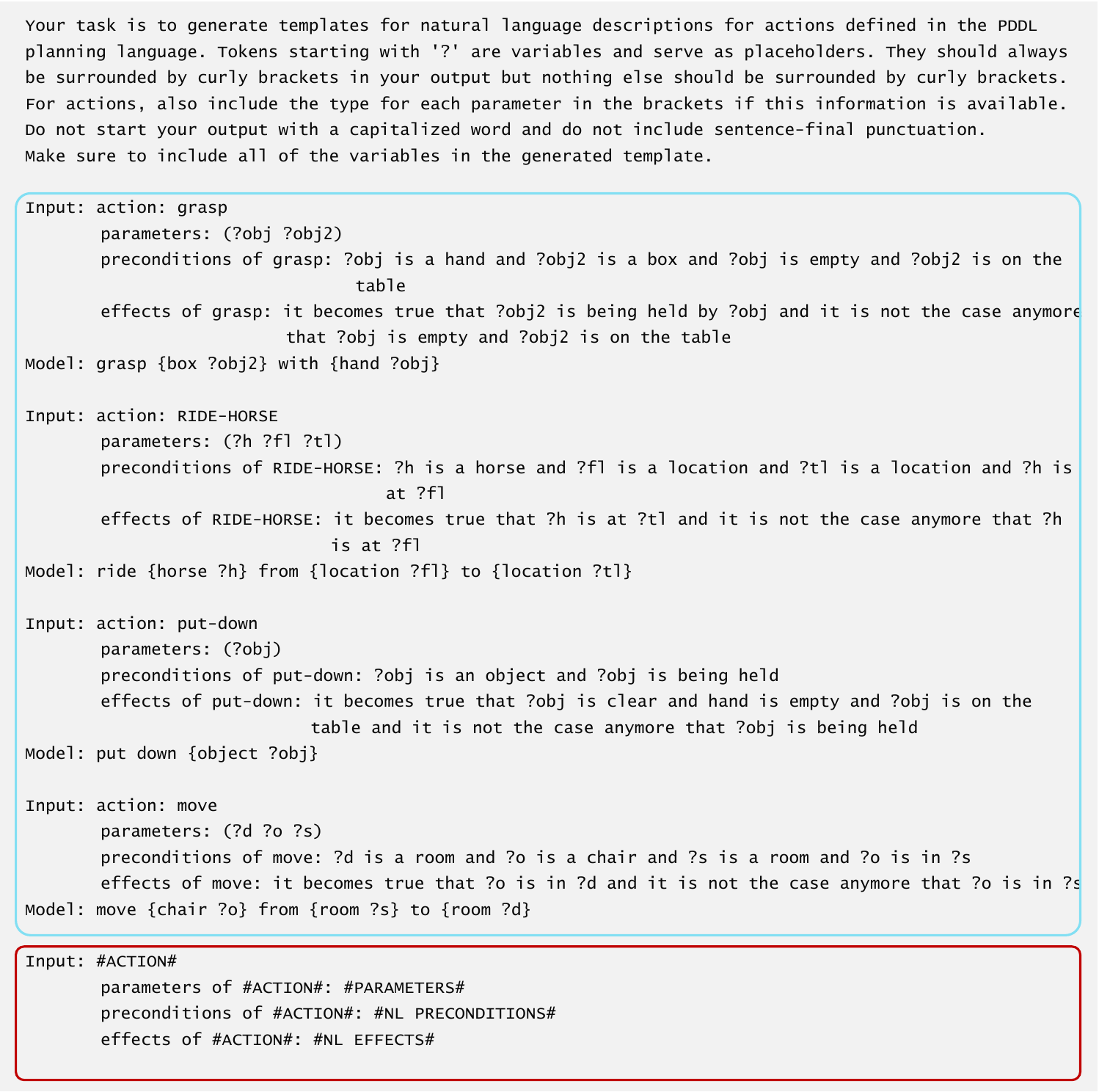}
    \caption{The complete prompt for the \APBLLM, used to generate the NL snippets for the PDDL actions.}
    \label{fig:apb_prompt_act}
\end{figure*}

%% file: appendix/4_planning_prompts.tex
\begin{figure*}[ht]
    \centering
    \includegraphics[width=0.75\textwidth]{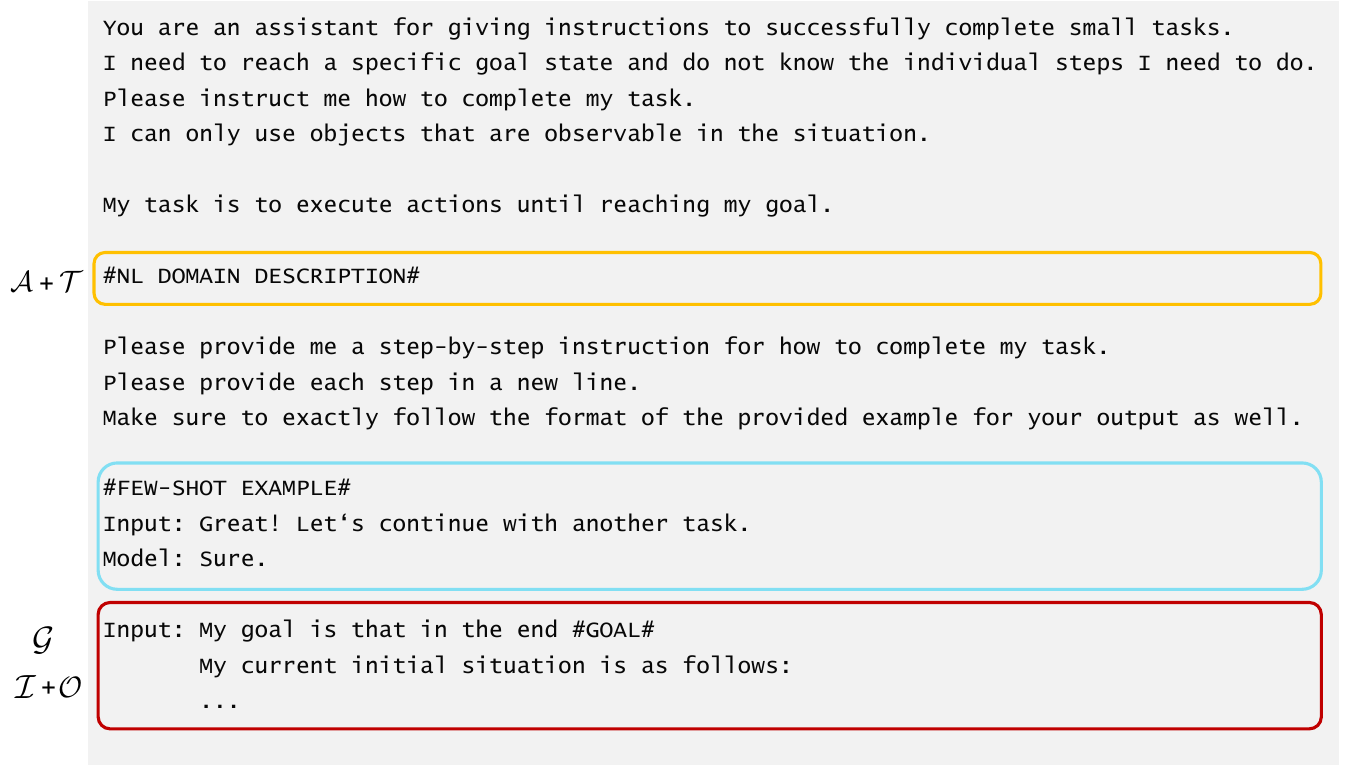}
    \caption{Structure of the prompt for the \PLLM\ used for the \textbf{Basic} planning mechanism. \#NL DOMAIN DESCRIPTION\# corresponds to the domain description and the \#FEW SHOT EXAMPLE\# is the approach-specific few-shot example. At the bottom, the beginning of the target problem description is shown.}
    \label{fig:P-LLM_prompt_basic}
\end{figure*}

\begin{figure*}[ht]
    \centering
    \includegraphics[width=0.75\textwidth]{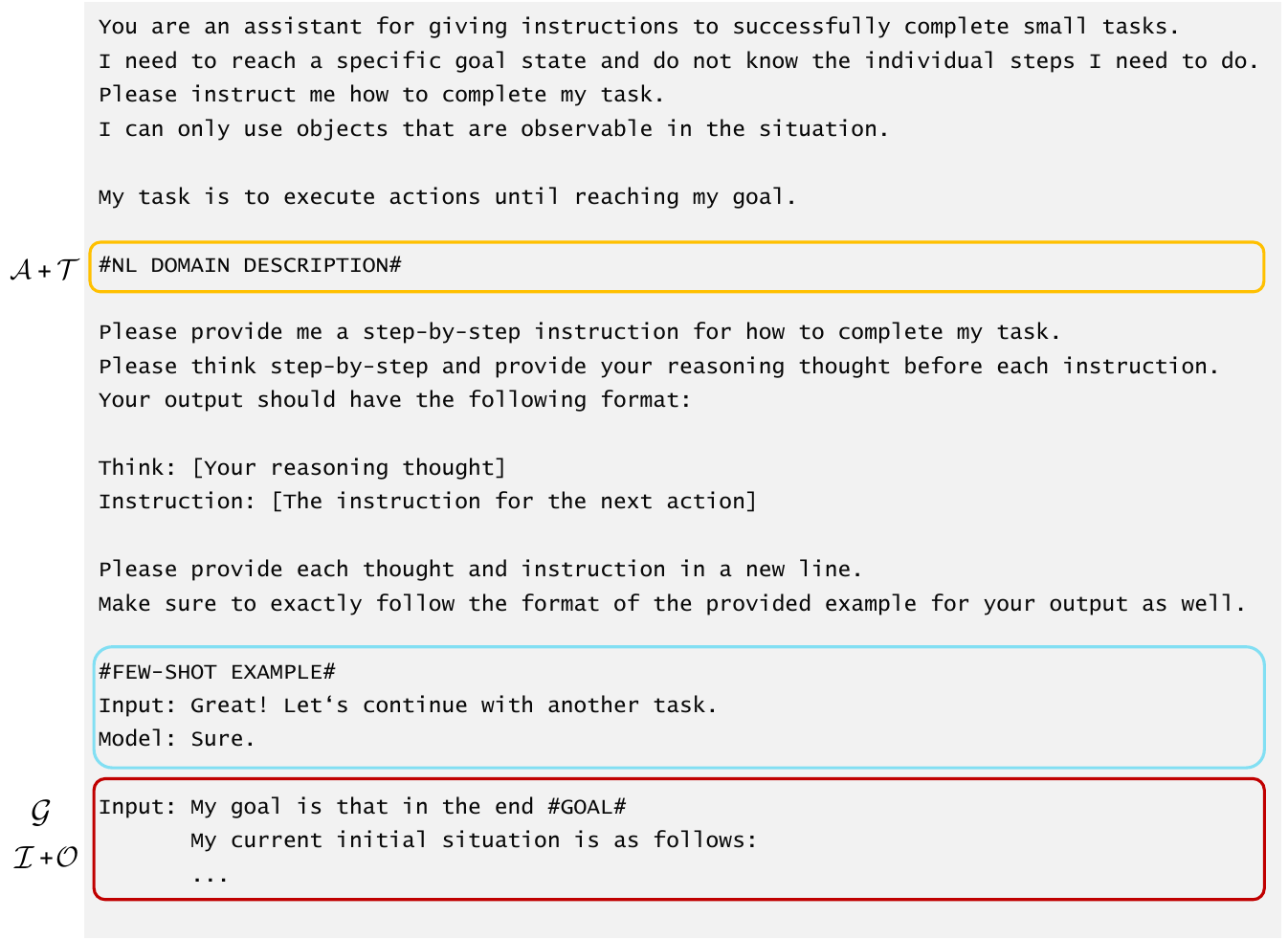}
    \caption{Structure of the prompt for the \PLLM\ used for the \textbf{CoT} planning mechanism. \#NL DOMAIN DESCRIPTION\# corresponds to the domain description and the \#FEW SHOT EXAMPLE\# is the approach-specific few-shot example. At the bottom, the beginning of the target problem description is shown.}
    \label{fig:P-LLM_prompt_cot}
\end{figure*}

\begin{figure*}[ht]
    \centering
    \includegraphics[width=0.75\textwidth]{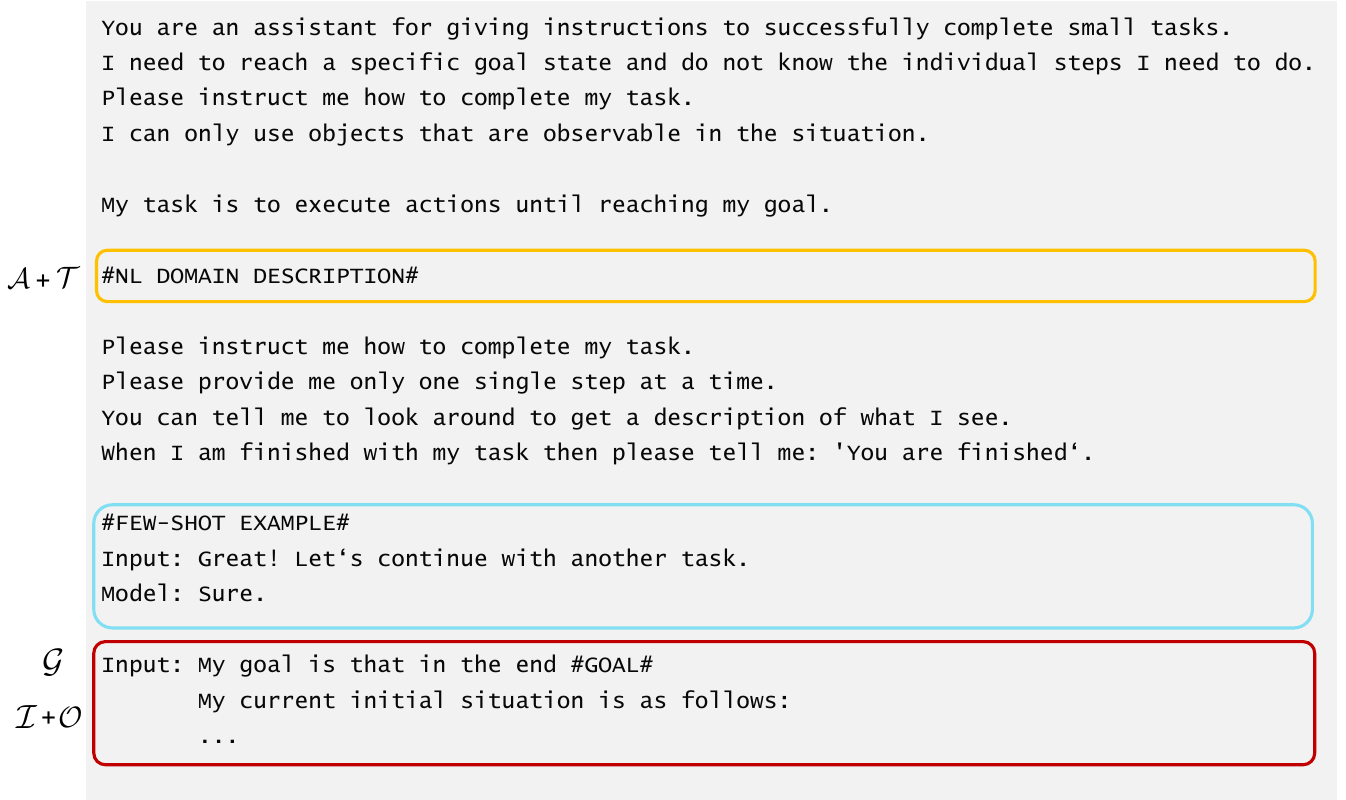}
    \caption{Structure of the prompt for the \PLLM\ used for the \textbf{Act} planning mechanism. \#NL DOMAIN DESCRIPTION\# corresponds to the domain description and the \#FEW SHOT EXAMPLE\# is the approach-specific few-shot example. At the bottom, the beginning of the target problem description is shown.}
    \label{fig:P-LLM_prompt_act}
\end{figure*}

\begin{figure*}[ht]
    \centering
    \includegraphics[width=0.75\textwidth]{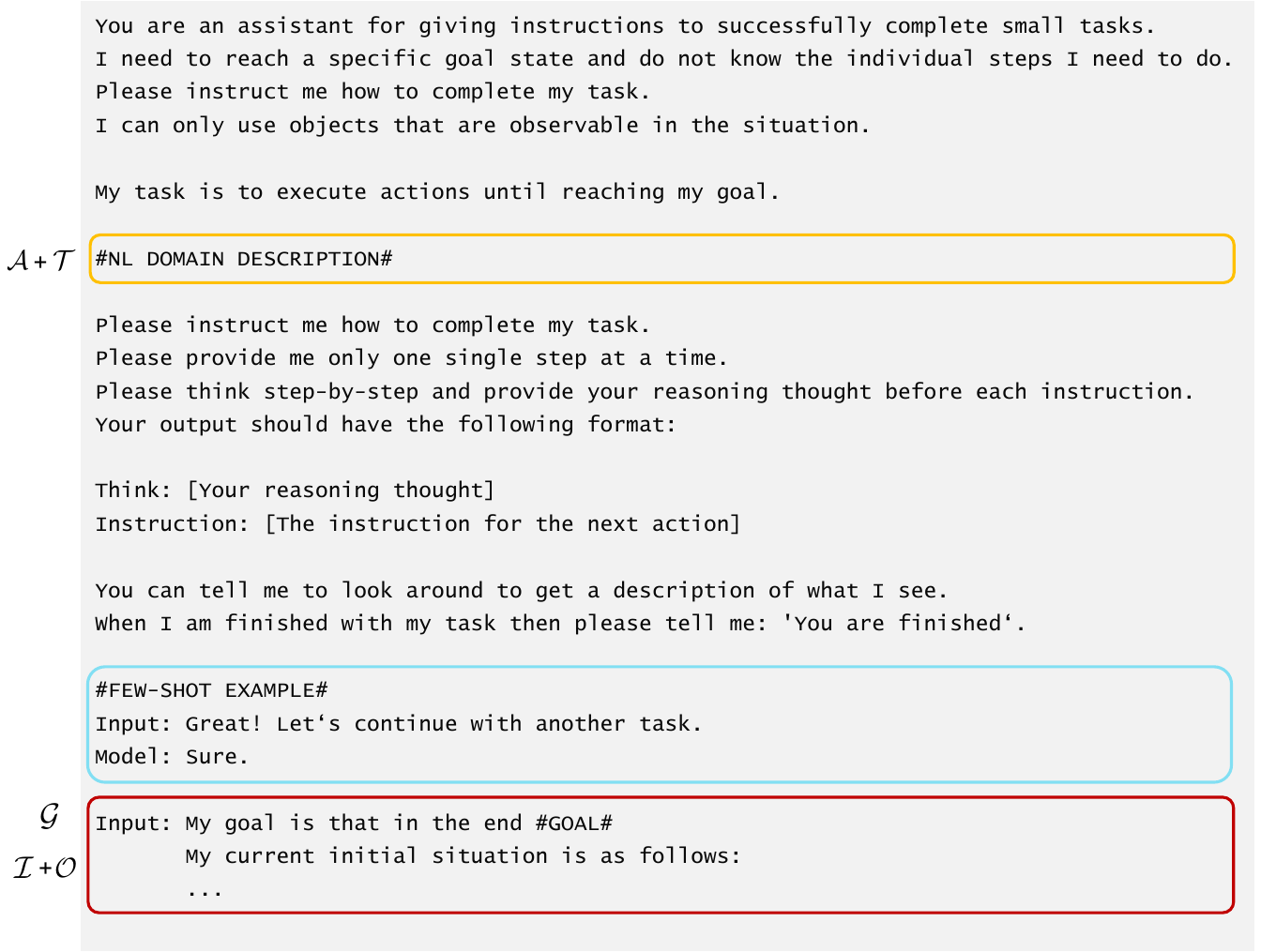}
    \caption{Structure of the prompt for the \PLLM\ used for the \textbf{ReAct} planning mechanism. \#NL DOMAIN DESCRIPTION\# corresponds to the domain description and the \#FEW SHOT EXAMPLE\# is the approach-specific few-shot example. At the bottom, the beginning of the target problem description is shown.}
    \label{fig:P-LLM_prompt_react}
\end{figure*}

\begin{figure*}
    \centering
    \includegraphics[width=0.7\textwidth]{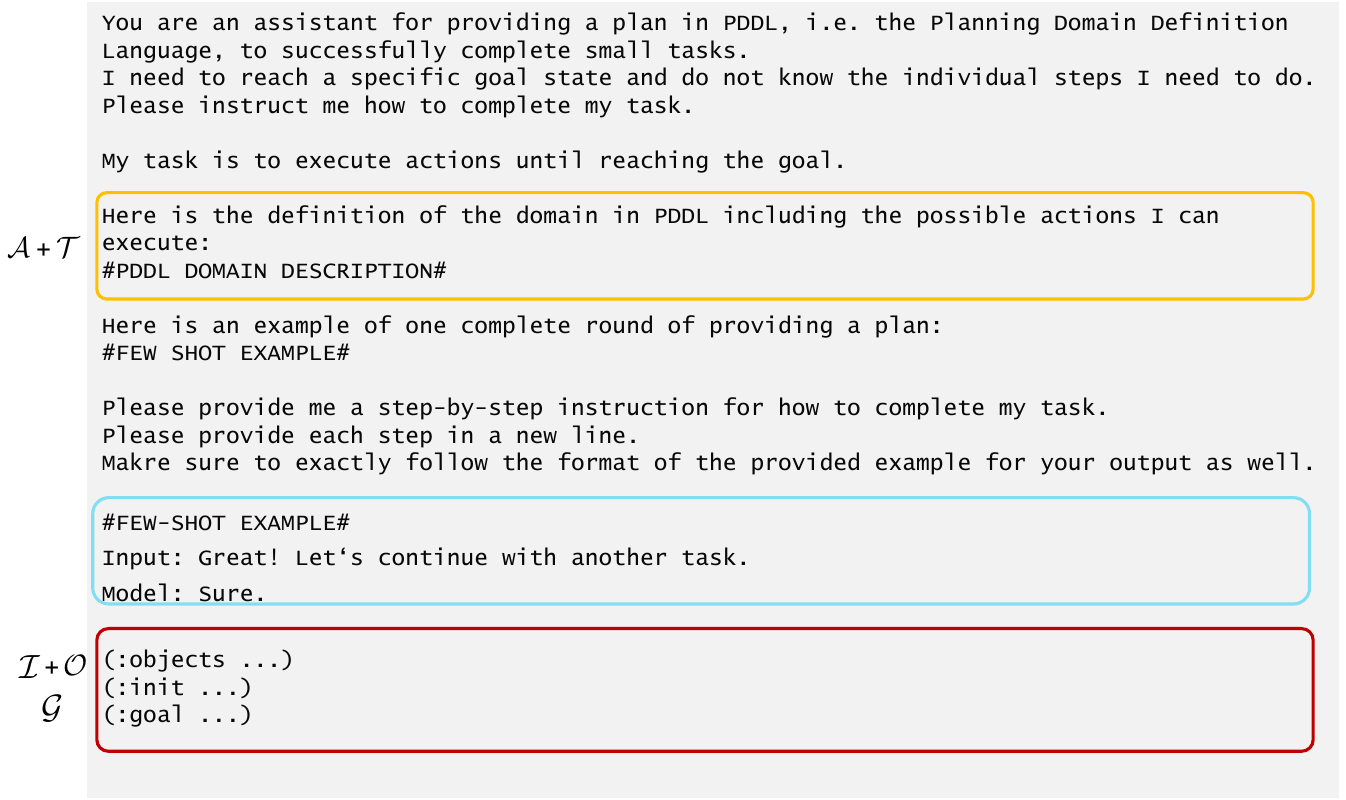}
    \caption{Structure of the prompt for the \PLLM\ used for the \textbf{Basic} LLM planning mechanism with PDDL as input. }
    \label{fig:prompt_basicpddl_struct}
\end{figure*}

\begin{figure*}
    \centering
    \includegraphics[width=0.7\textwidth]{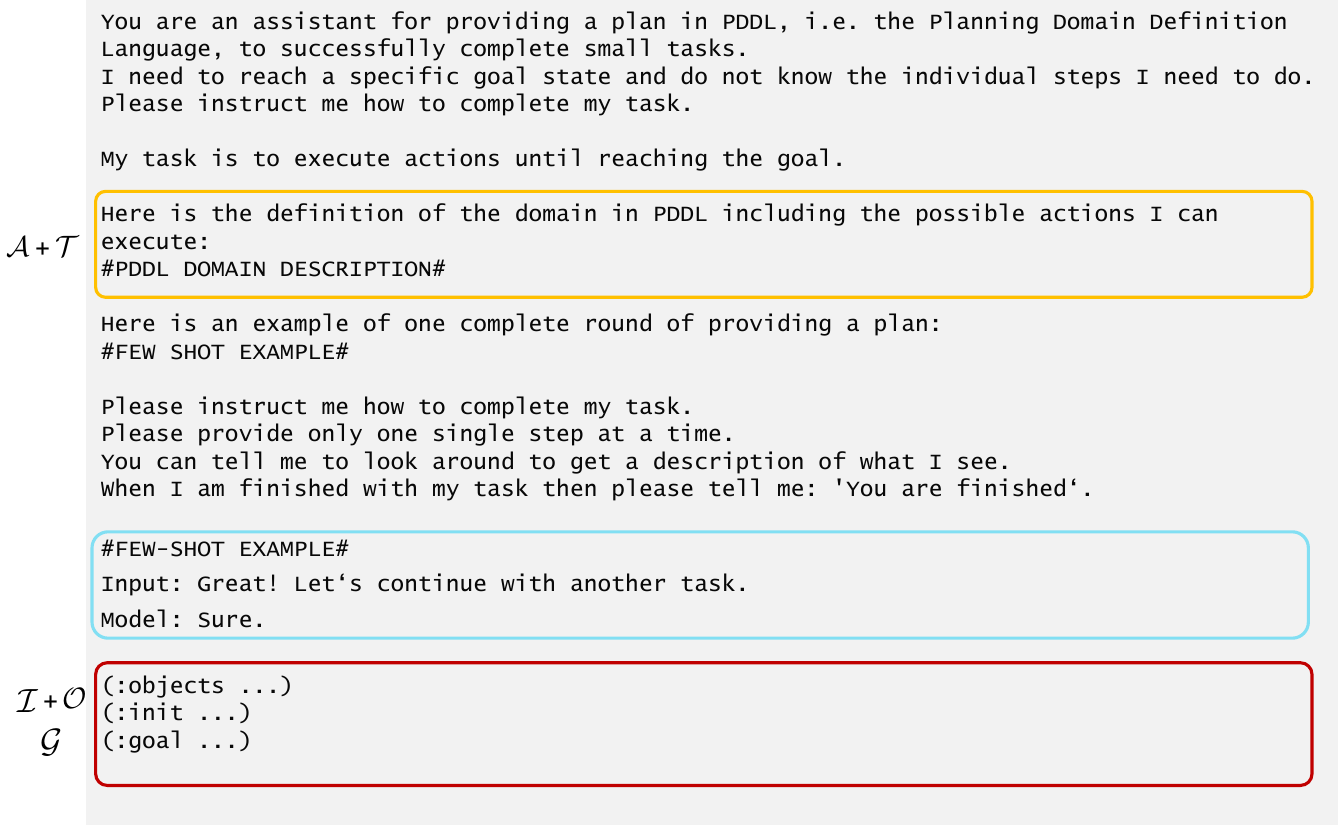}
    \caption{Structure of the prompt for the \PLLM\ used for the \textbf{Act} LLM planning mechanism with PDDL as input. }
    \label{fig:prompt_actpddl_struct}
\end{figure*}

\begin{figure*}[ht]
    \centering
    \includegraphics[width=0.7\textwidth]{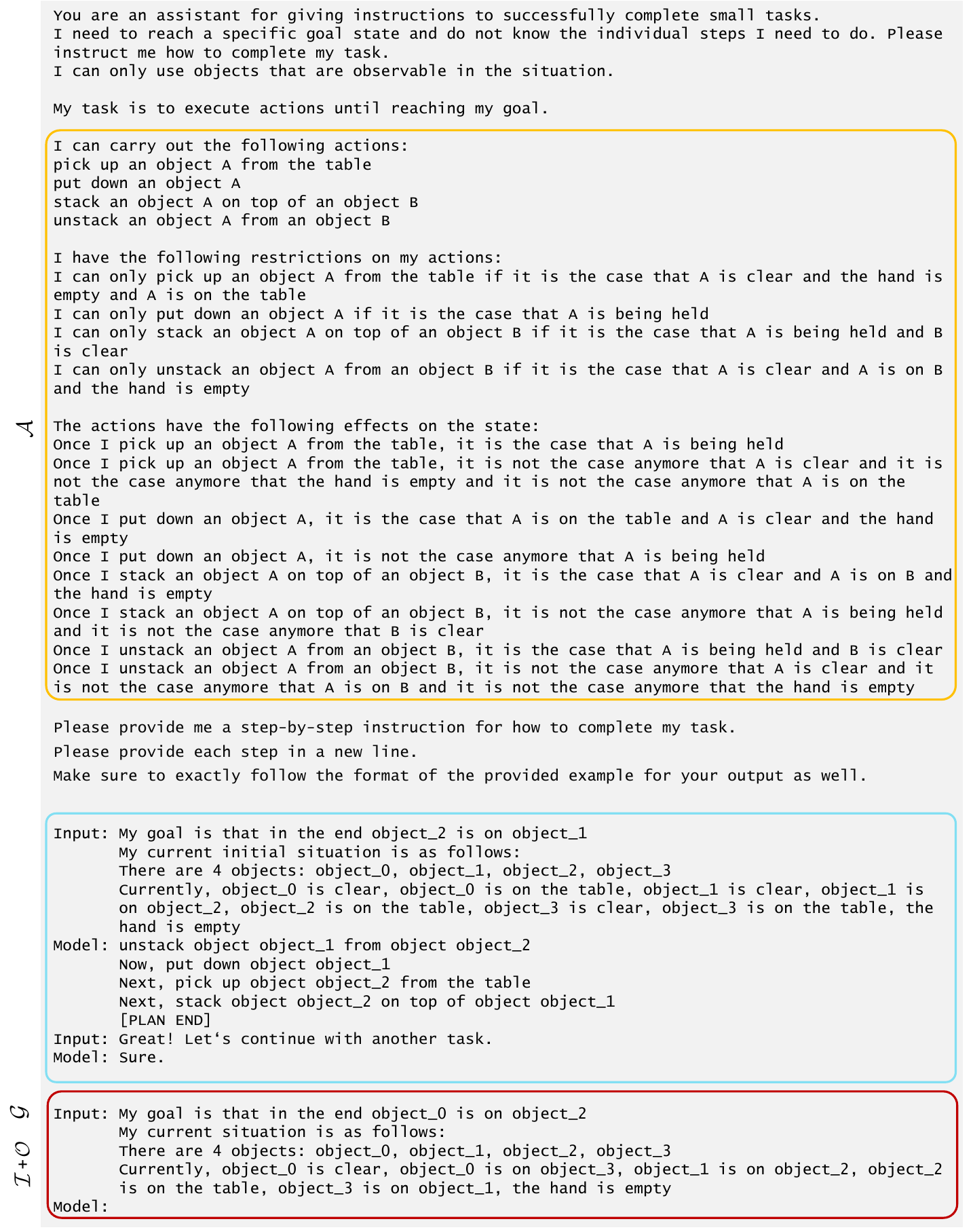}
    \caption{Example of a complete input prompt for the \textbf{Basic} action-choice mechanism for a problem from the the Blocksworld domain consisting of the overall instructions for the LLM, the Blocksworld NL domain description and one few-shot example followed by the description of the target problem.}
    \label{fig:prompt_ex_basic}
\end{figure*}

\begin{figure*}[ht]
    \centering
    \includegraphics[width=0.7\textwidth]{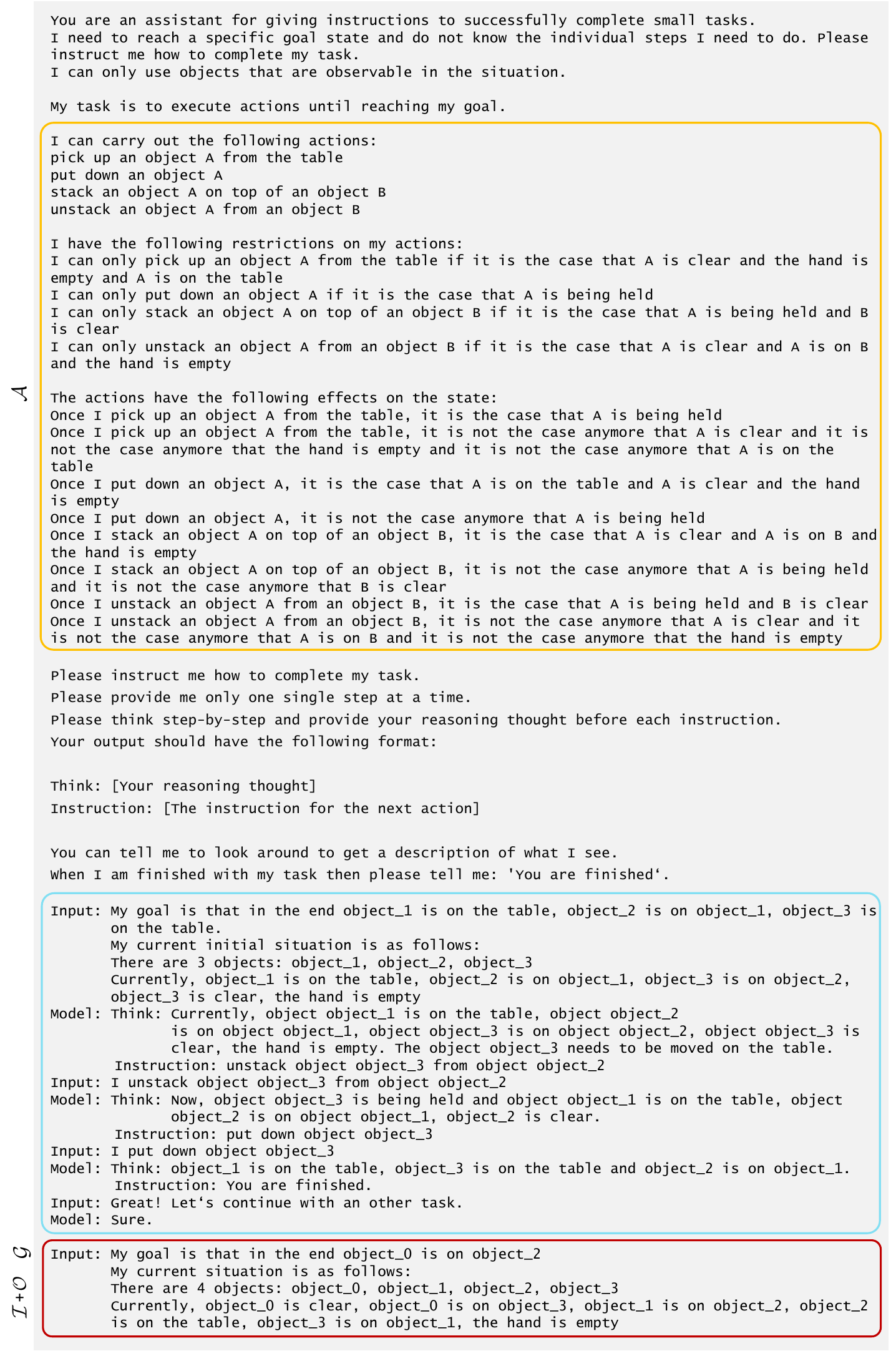}
    \caption{Example of a complete input prompt for the \textbf{ReAct} action-choice mechanism for a problem from the the Blocksworld domain consisting of the overall instructions for the LLM, the Blocksworld NL domain description and one few-shot example followed by the description of the target problem.}
    \label{fig:prompt_ex_react}
\end{figure*}

\begin{figure*}
    [ht]
    \centering
    \includegraphics[width=0.7\textwidth]{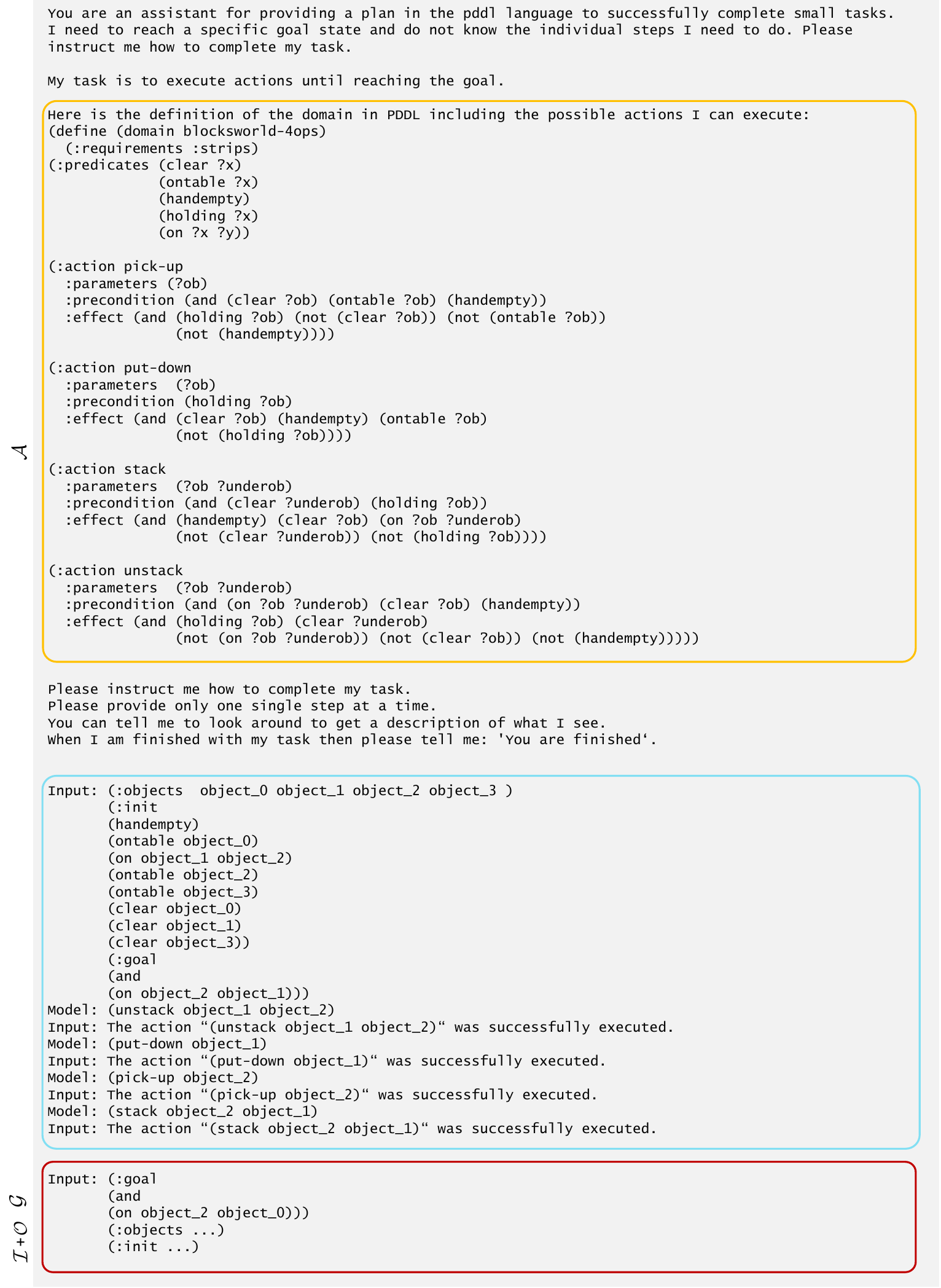}
    \caption{Example of a complete input prompt for the \textbf{Act} action-choice mechanism with PDDL input for a problem from the the Blocksworld domain consisting of the overall instruction for the LLM, the Blocksworld domain description and one few-shot example followed by the description of the target problem.}
    \label{fig:prompt_ex_pddl_act}
    
\end{figure*}

\begin{figure*}[ht]
    \centering
    \includegraphics[width=0.7\textwidth]{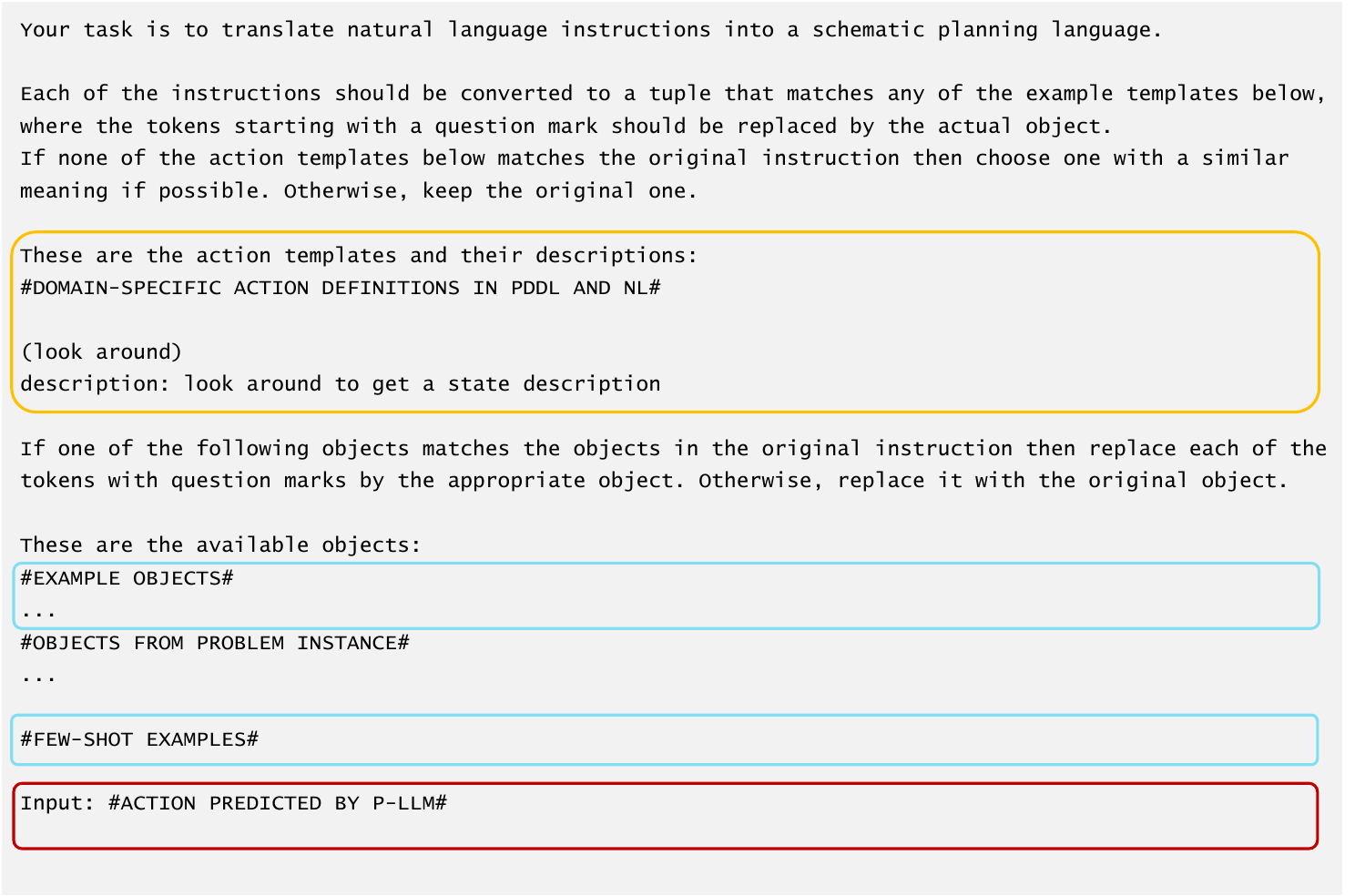}
    \caption{The structure of the prompt for the \TLLM\ to translate the predicted NL actions back into PDDL. The prompt consists of the general task instruction at the beginning followed by the domain-specific pairs of PDDL and NL action descriptions. All available objects are included in the prompt as well as some example objects that are used in the few-shot examples provided.}
    \label{fig:t_llm_prompt_template}
\end{figure*}

\begin{figure*}[ht]
        \centering
    \includegraphics[width=0.7\textwidth]{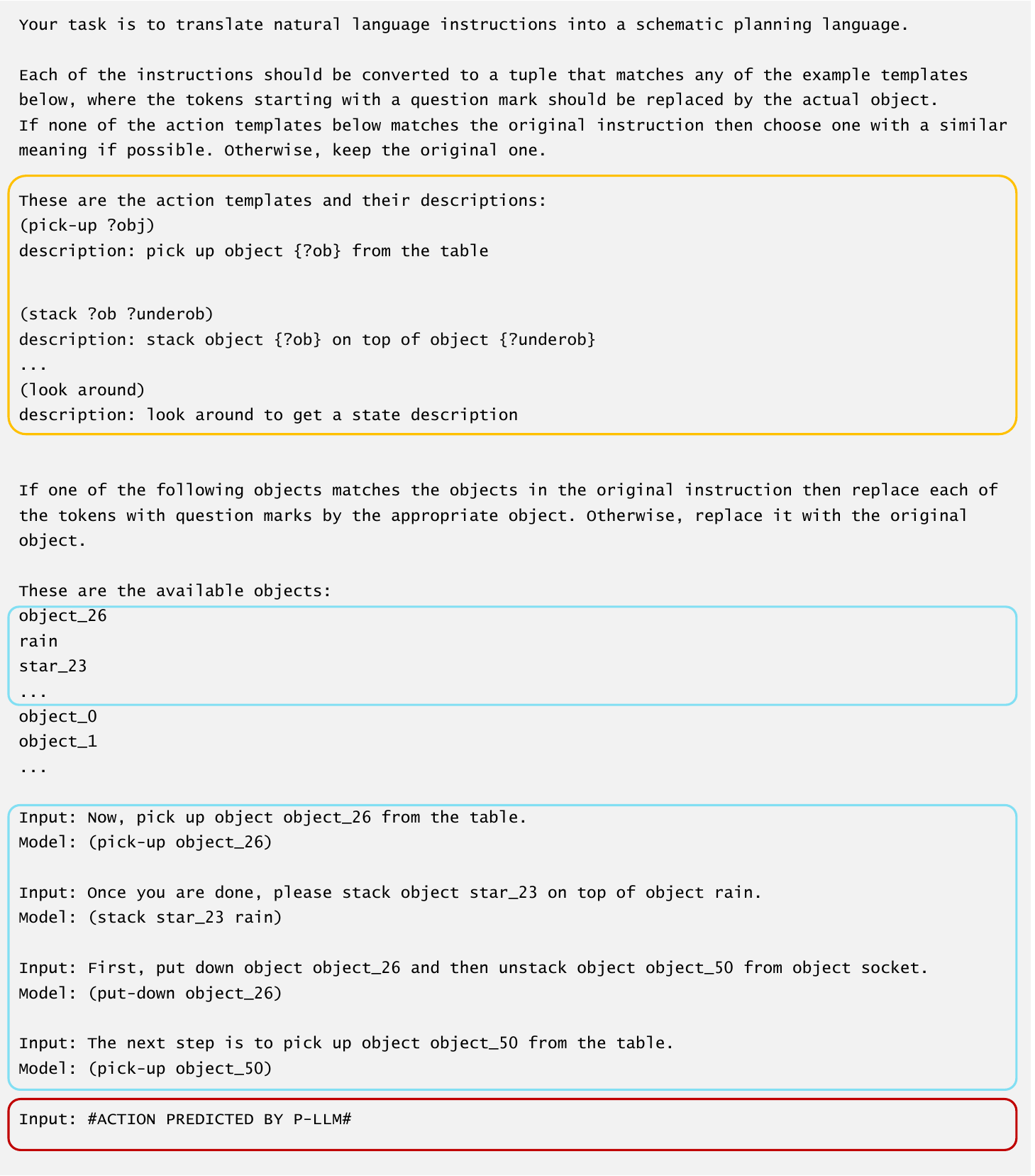}
    \caption{The prompt for the \TLLM\ to translate the predicted NL actions from the Blocksworld domain back into PDDL.}
    \label{fig:t_llm_prompt}
\end{figure*}

%% file: appendix/5_react_few_shot.tex
\begin{figure*}
    \centering
    \includegraphics[width=0.75\textwidth]{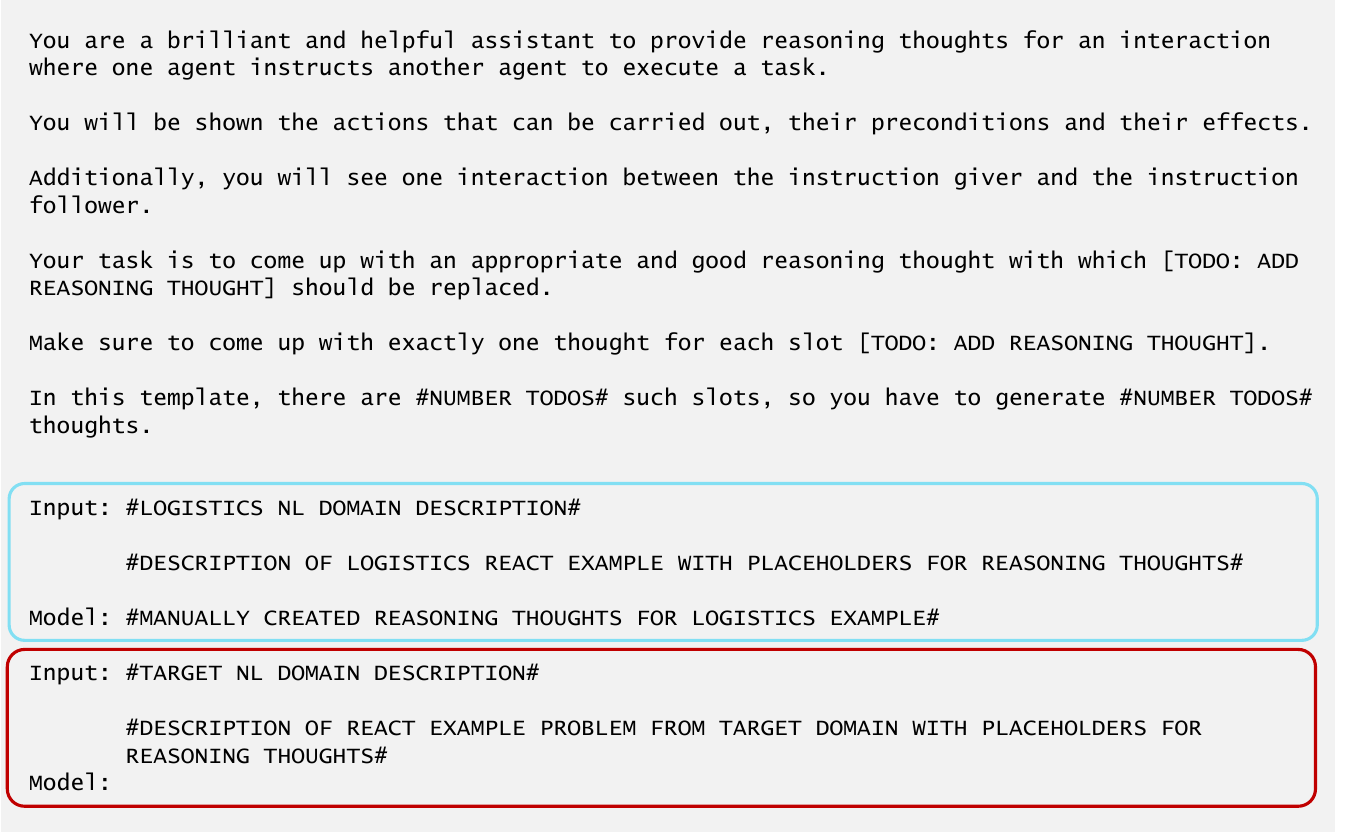}
    \caption{Structure of the prompt for generating the thoughts for the \texttt{ReAct} and \texttt{CoT} few-shot examples using an LLM. The LLM is provided a general task instruction followed by the manually created Logistics few-shot example and the domain description and gold plan trajectory of the target problem.}
    \label{fig:thought_gen_temp}
\end{figure*}

\begin{figure*}
    \centering
    \includegraphics[width=0.75\textwidth]{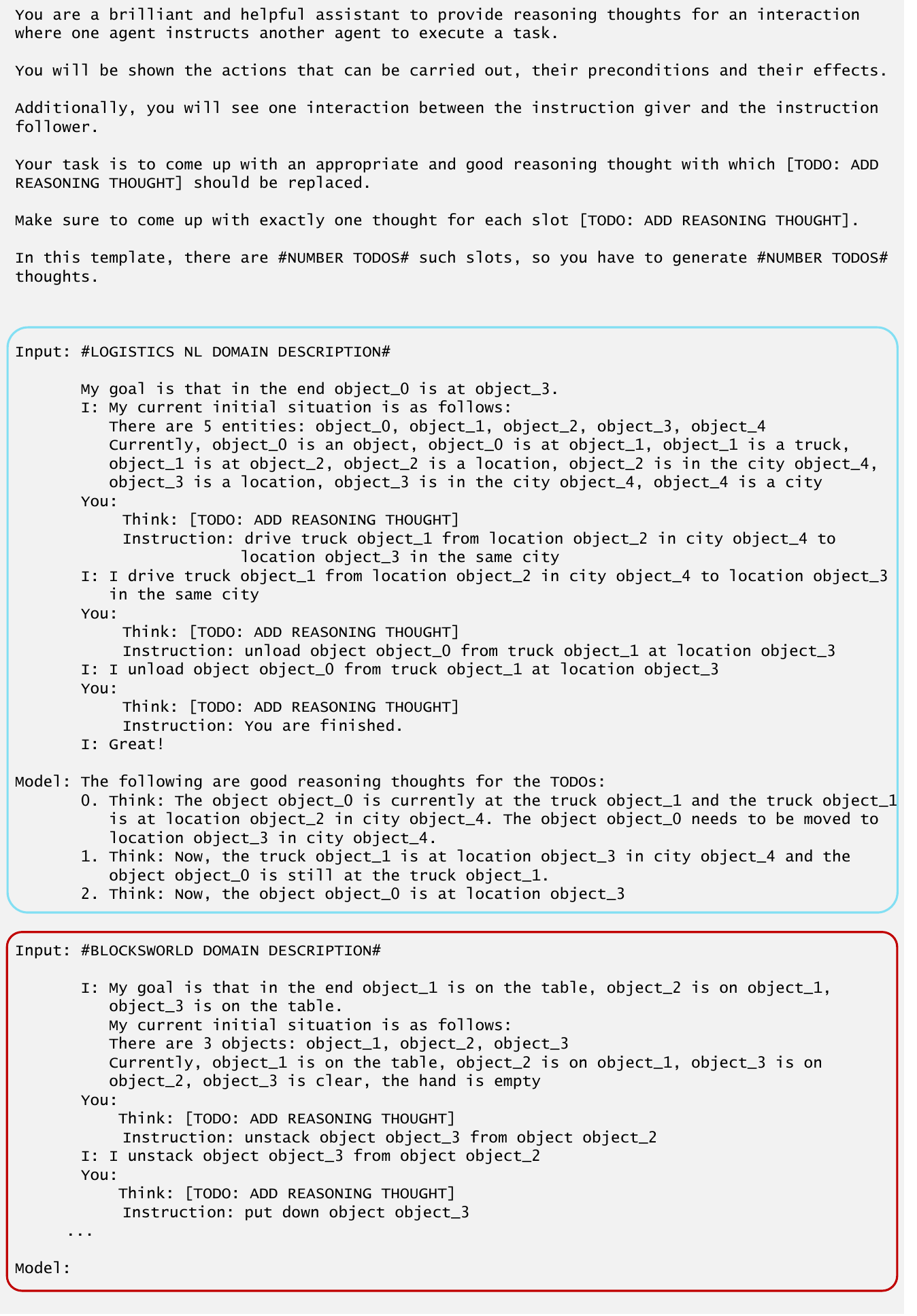}
    \caption{Example of the prompt for generating the thoughts for the \texttt{ReAct} and \texttt{CoT} few-shot examples for a Blocksworld problem based on the manually created example from the Logistics domain.}
    \label{fig:thought_gen1}
\end{figure*}

